%% file: SMC2023.tex
\documentclass[letterpaper, 10 pt, conference]{ieeeconf}  

\IEEEoverridecommandlockouts                              

\usepackage{cite}
\usepackage{amsmath,amssymb,amsfonts}
\usepackage{algorithmic}
\usepackage{graphicx}
\usepackage{textcomp}
\usepackage{xcolor}
\usepackage{printlen}

\usepackage{amsmath} 
\usepackage{amssymb}  
\usepackage{xcolor}
\usepackage{svg}
\usepackage{pgf}
\usepackage{tikz}
\usepackage{textcomp}
\usepackage{hyperref}
\usepackage{lipsum}

\usepackage[normalem]{ulem}

\newcommand\copyrighttext{%
  \footnotesize \textcopyright 2023 IEEE. Personal use of this material is permitted.
  Permission from IEEE must be obtained for all other uses, in any current or future 
  media, including reprinting/republishing this material for advertising or promotional 
  purposes, creating new collective works, for resale or redistribution to servers or 
  lists, or reuse of any copyrighted component of this work in other works. 
  }
\newcommand\copyrightnotice{%
\begin{tikzpicture}[remember picture,overlay]
\node[anchor=south,yshift=10pt] at (current page.south) {\fbox{\parbox{\dimexpr\textwidth-\fboxsep-\fboxrule\relax}{\copyrighttext}}};
\end{tikzpicture}%
}

\title{\LARGE \bf
Shared telemanipulation with VR controllers in an anti slosh scenario*
}

\author{Max Grobbel$^{1}$ and Balint Varga$^{2}$ and Sören Hohmann$^{2}$
\thanks{*This work is partly supported by the Federal Ministry of Education and Research, in the START-interaktiv research initiative with the project number 16SV8801.}
\thanks{$^{1}$Max Grobbel is with FZI - Forschungszentrum Informatik, 76131 Karlsruhe, Germany
        {\tt\small grobbel@fzi.de}}%
\thanks{$^{2}$Balint Varga and Sören Hohmann are with the Institute of Control Systems, Karlsruhe Institute of Technology,
        76131 Karlsruhe, Germany
        {\tt\small balint.varga2@kit.edu} and {\tt\small soeren.hohman@kit.edu}}%
}

\def\BibTeX{{\rm B\kern-.05em{\sc i\kern-.025em b}\kern-.08em
    T\kern-.1667em\lower.7ex\hbox{E}\kern-.125emX}}

\begin{document}

\maketitle
\copyrightnotice
\thispagestyle{empty}
\pagestyle{empty}

\begin{abstract}
Telemanipulation has become a promising technology that combines human intelligence with robotic capabilities to perform tasks remotely. However, it faces several challenges such as insufficient transparency, low immersion, and limited feedback to the human operator. Moreover, the high cost of haptic interfaces is a major limitation for the application of telemanipulation in various fields, including elder care, where our research is focused. \textbf{To address these challenges, this paper proposes the usage of nonlinear model predictive control for telemanipulation using low-cost virtual reality controllers, including multiple control goals in the objective function}. The framework utilizes models for human input prediction and task-related models of the robot and the environment.  The proposed framework is validated on an UR5e robot arm in the scenario of handling liquid without spilling. Further extensions of the framework such as pouring assistance and collision avoidance can easily be included.
\end{abstract}


\input{01_Introduction}

\input{02_Related_Work}
\input{03_Models}

\input{04_Framework}
\input{05_Experiments}

\input{06_Conclusion}




\bibliographystyle{IEEEtran}
\bibliography{SMC}

\vspace{20pt}

\end{document}

%% file: 01_Introduction.tex
\section{Introduction}
Telemanipulation is an emerging field that aims to combine the skills of human operators with robotic systems to perform tasks in a variety of fields \cite{Sheridan.2016}. These fields include elder care, handling hazardous materials, and space exploration. In particular, telemanipulation has the potential to overcome a number of challenges, including inaccessibility in hazardous environments, lack of human resources, and the need for precision in certain applications.

However, the success of telemanipulation is hindered by several challenges \cite{Hokayem.2006}. The two main challenges are the communication delays and the high cost of haptic interfaces of the state-of-the-art applications. The reasons of such communication delays and dropouts are network latency, packet loss, or hardware failure, which can result in unpredictable and unstable behavior of the remote robot. Haptic interfaces are essential in order to provide feedback to the operator to enhance the intuition of the control \cite{Hirche.2012, Wildenbeest.2013}. Since, such haptic interfaces are costly, the application of telemanipulation in various fields are limited due to this cost factor.

To address these challenges, this paper proposes a novel framework for telemanipulation using virtual reality (VR) controllers and nonlinear model predictive control (NMPC). The framework aims to provide efficient models for predicting human inputs and enabling the efficient execution of the remote tasks even in the presence of communication issues. The use of VR controllers provides a low-cost alternative to traditional haptic interfaces while maintaining the operator's intuitive control of the remote robot. Additionally, the NMPC algorithm improves the overall stability and accuracy of the telemanipulation system.

The proposed framework is validated on an UR5e robot arm with a glass of water connected to the end effector. The framework provides an anti-slosh assistance for the handling of liquid containers, which is a challenging use case due to the sloshing dynamics of the liquid. 

The remainder of this paper is structured as follows: In section \ref{sec:relatedWork} we give an overview of related literature. The description of utilized system models is given in section \ref{section:models}. Our proposed Assistive Telemanipulation Framework is presented in section \ref{sec:Framework}. In section \ref{sec:Realization} we show the realization of the framework on a real UR5e robot.

%% file: 02_Related_Work.tex
\section{Related Work} \label{sec:relatedWork}
In the following, we give a brief overview of related research in the fields of bilateral telemanipulation, model predictive control in the context of robotics and anti slosh control.

\subsection{Bilateral Telemanipulation}
Literature in the field of telemanipulation mainly focuses on setups with haptic input devices, also known as bilateral telemanipulation \cite{Hokayem.2006}, where the human operator also receives haptic feedback and becomes part of the control loop. The goal of those approaches is to support the human with a transparent telemanipulation system \cite{Hirche.2012}, \cite{Wildenbeest.2013} and high immersion \cite{Hertkorn.2016}, \cite{Nostadt.2020} utilizing the haptic feedback.
A disadvantage of bilateral telemanipulation is the high cost of the input devices (e.g. \cite{sigma7}).




\subsection{Robotics and Model Predictive Control}
Model predictive control (MPC) utilizes model knowledge and an objective function to calculate optimal trajectories that satisfy the system dynamics. One challenge with MPC is the real-time capability. With increasing calculation power of modern processors and efficient solvers for optimization problems, MPCs are applied more and more in robotic applications, e.g. in \cite{Fau:17} frequencies of $1 \mathrm{kHz}$ are accomplished. Their robot model is given in joint space, such that the inverse kinematics are solved inherently by the optimization. In telemanipulation robotics, MPCs can be utilized for collision avoidance, though they often are combined with haptic input devices \cite{Hu.2021}. In \cite{Rubagotti.2019} the implementation of a MPC for collision avoidance on a telemanipulated UR5e is presented. To deal with the high calculation times of online optimization problems, approximate MPCs based on Neural Networks are suggested in \cite{Nubert.2020}.

\subsection{Anti Slosh Control}
When transporting liquids in an open container, the sloshing dynamics of the liquid have to be considered. In \cite{Guagliumi.2021}, the modelling of the liquid either as a pendulum or as a mass, damper spring system is compared in different scenarios.
The examination of liquid in a hemispherical container with effective anti slosh control is presented in \cite{Feddema.1997}. Their algorithm is based on a model of the liquid as a pendulum. The control architecture in \cite{Biagiotti.2018} is also based on the modelling as a pendulum and implements input filter to generate slosh free movements in telemanipulation. It is worth mentioning that this work does not rely on haptic input devices, but uses motion detection with cameras as human input interface. Based on the pendulum model, \cite{ReinholdJan.2019} and \cite{Maderna.2018} generate optimal trajectories which are used as references for the robot controller. The work of \cite{Muchacho.2022} is not based on a liquid model directly, but they enforce trajectories of the telemanipulated robot arm, such that the container does not experience any lateral acceleration.

So far, no telemanipulation framework based on MPC for anti slosh control with VR controllers as input devices has been proposed in literature.

%% file: 03_Models.tex
\section{Models for Assistive Telemanipulation Framework} \label{section:models}
The assistive telemanipulation framework is based on two underlying system models, namely the model of the robot arm and the model of liquid in a container, which are derived in this section.

\subsection{Model of Robot}

\begin{figure}
\begin{center}
\includegraphics[width=8.4cm]{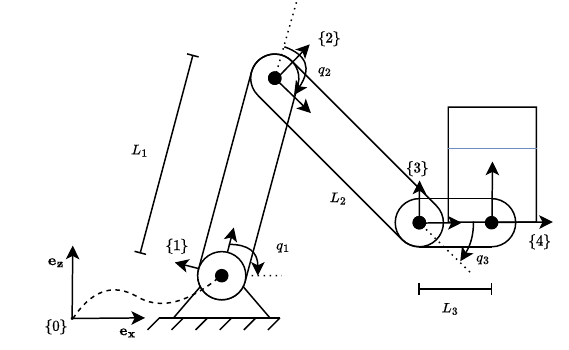}    
\caption{Planar robot arm with three rotational joints $q_i$ and a glass of water with the coordinate frame $\{4\}$ connected to the end effector.} 
\label{fig:planar3R}
\end{center}
\end{figure}

A generic robot arm with $n_q$ rotational joints can be described through the states $[\mathbf{q}^T, \mathbf{\dot{q}}^T]^T$, where $\mathbf{q} \in \mathbb{R}^{n_q}$ denotes all joint angles and $\mathbf{\dot{q}} \in \mathbb{R}^{n_q}$ the joint angular velocities. The relation of the position and orientation of the end effector in global cartesian coordinates $\mathbf{x}$ and the robot joints  $\mathbf{q}$ is given through the forward kinematics $F_K:\mathbb{R}^{n_q} \rightarrow \mathbb{R}^{7}, \mathbf{q} \mapsto \mathbf{x}$. $\mathbf{x}$ consists of the cartesian coordinates of the end effector $\mathbf{r}$ and the rotation described as an union quaternion. 

The dynamics of a robot are described through the set of second order differential equations of the form \cite{Lynch.2017}
\begin{equation} \label{eq:RoboterDynamik}
    \mathbf{M}(\mathbf{q}) \mathbf{\ddot{q}} + \mathbf{b}(\mathbf{q},\mathbf{\dot{q}}) = \mathbf{\tau},
\end{equation}
where $\mathbf{\tau}$ contains the applied torques and $\mathbf{\ddot{q}}$ the acceleration in all joints, whereas $\mathbf{M}(\mathbf{q})$ 
denotes the inertia tensor and $\mathbf{b}(\mathbf{q},\mathbf{\dot{q})}$ all further relevant terms (dissipation, Coriolis effects, gravity). 

Like \cite{Nubert.2020} and \cite{Fau:17}, we assume the existence of low-level controllers such that the joint accelerations serve as control inputs $\mathbf{u} = \mathbf{\ddot{q}}, \mathbf{u} \in \mathbb{R}^{n_q}$ of the considered system. With this assumption, the equations of motion (\ref{eq:RoboterDynamik}) are reduced to the linear dynamic system
\begin{equation} \label{eq:stateSpaceModelRobot}
    \frac{\mathrm{d}}{\mathrm{d}t}
    {\begin{bmatrix}
    \mathbf{q}\\
    \mathbf{{\dot{q}}}
    \end{bmatrix}}
 = \mathbf{A} \begin{bmatrix}
    \mathbf{q}\\ \mathbf{\dot{q}}
    \end{bmatrix} + \mathbf{B} \mathbf{u}
\end{equation}
with 
\begin{align}\nonumber
\mathbf{A} = \begin{bmatrix} \mathbf{0} && \mathbf{\mathbb{I}}\\\mathbf{0} && \mathbf{0} \end{bmatrix}
\qquad \mathrm{and} \qquad
\mathbf{B} = \begin{bmatrix} \mathbf{0} \\ \mathbf{\mathbb{I}} \end{bmatrix},
\end{align}
where $\mathbb{I} \in \mathbb{R}^{n_q \times n_q}$ denotes the identity matrix and $\mathbf{0} \in \mathbb{R}^{n_q \times n_q}$.

For the evaluation, an UR5e robot with 6 degrees of freedom is utilized. Similar to \cite{Fau:17}, only the joints {${\{q_1,q_2,q_3\} \subset \{q_0,q_1,q_2,q_3,q_4,q_5\}}$} are actuated, thus only pure planar movement is considered in this work as shown in Fig.~\ref{fig:planar3R}. The angles $q_1,q_2$ and $q_3$ are depicted in positive orientation. The coordinate frames $\{1\},\{2\}$ and $\{3\}$ are connected to the robot links with length $L_1, L_2$ and $L_3$, respectively. A glass $\{4\}$ with water is connected to the end effector. The world frame $\{0\}$ has the same origin as frame $\{1\}$.

With $n_q=3$ the system, states reduce to $[\mathbf{q}^T, \mathbf{\dot{q}}^T]^T = [q_1, q_2, q_3, \dot{q}_1, \dot{q}_2, \dot{q}_3 ]^T$. With the lengths $L_1, L_2$ and $L_3$ of the robot links, the forward kinematics for this planar configuration 
$F_{K,2D}:\mathbb{R}^{3} \rightarrow \mathbb{R}^3, \mathbf{q} \mapsto   [x_c,z_c,\theta_c]^T$ are described by \cite[p.137]{Lynch.2017}
\begin{subequations}
\label{eq:ForwardKinematics}
\begin{align}
    x_c &= L_1\cos q_1 + L_2 \cos(q_1+q_2)\\ \nonumber
            &+ L_3 \cos(q_1 + q_2 + q_3),\\
    z_c &= -( L_1\sin (q_1) + L_2 \sin(q_1 + q_2) \\ \nonumber
            &+ L_3 \sin(q_1 + q_2 + q_3)), \\ 
    \theta_c &= q_1 + q_2 + q_3.
\end{align}
\end{subequations}

\subsection{Model of Liquid in Open Container}

\begin{figure}
\begin{center}
\includegraphics[width=\linewidth]{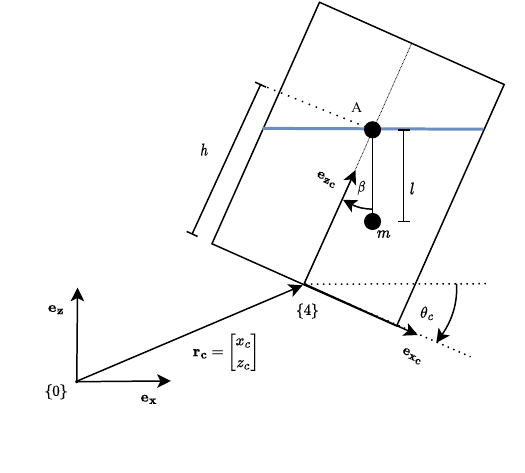}    
\caption{The liquid in a container is modelled as a pendulum.} 
\label{fig:FluidModel}
\end{center}
\end{figure}

The transportation of liquids in open containers is discussed in \cite{Feddema.1997}, which is adapted for our framework. Under the assumption, that the surface of the liquid stays flat during movements, it can be modeled as an oscillating pendulum with a moving base \cite{Feddema.1997}. The revolution point A of the pendulum lies on the intersection of the surface of the liquid and the middle line of the container and the deflection $\beta$ is measured relative to the $\mathbf{e}_z$ axis of the glass $\{4\}$ (Fig. \ref{fig:FluidModel}). The surface is always perpendicular to the pendulum rod. Further, it is assumed that the intersection of the container is circular with a constant diameter.

The equation of movement of the pendulum is derived using the Lagrange Formalism. The position of the pendulum mass in the global coordinate frame $\{0\}$ is determined by
\begin{subequations}
\begin{align}
    &x_m = x_c + h \sin (\theta) - l \sin (\theta + \beta) \\
    &z_m = z_c + h \cos (\theta) - l \cos (\theta + \beta)
\end{align}
\end{subequations}
with the filling level $h$ of the liquid and the virtual pendulum length $l$. 
The pendulum length is a function of the viscosity of the liquid, diameter of the container and gravitation $g$ and can be determined by measuring the natural frequency of the liquid $\omega$ and the relation (see e.g.~\cite{Biagiotti.2018})
\begin{equation}
     \omega = \sqrt{\frac{g}{l}}. 
\end{equation}
Assuming that the movement of the container is enforced by the robot, the pendulum angle $\beta$ remains the only degree of freedom of the liquid subsystem. Thus, it is used as the generalized coordinate for the Lagrange Formalism. Introducing the Rayleigh dissipation function \cite{Feddema.1997}
\begin{equation}
    \mathcal{R} = \frac{1}{2}d\dot{\beta}^2
\end{equation}
with the damping coefficient $d$, the dynamics of the planar pendulum in the container of the end effector is expressed~as:
\begin{align} \label{eq:Bewegungsgleichung}
\begin{split}
       \ddot{\beta} &= f_{\beta}(\beta,\dot{\beta},\ddot{x}_c,\ddot{z}_c, \theta_c, \dot{\theta}_c, \ddot{\theta}_c) \\
       &=\frac{1}{l} \left( - (l - h \cos(\beta)) \ddot{\theta}_c + h\sin(\beta)\dot{\theta}_c^{2} \right. \\ 
      &+ \cos(\theta_c + \beta) \ddot{x}_{c} - \sin(\theta_c + \left.  \beta)(g+\ddot{z}_{c}) - \frac{d}{ml}\dot{\beta} \right), 
\end{split}
\end{align}
where the function $f_{\beta}$ describes the dependence of the pendulum's angular acceleration on the containers position, velocity, and the acceleration.
The dynamics of the liquid can thus be described through the nonlinear system
\begin{equation} \label{eq:stateSpaceModelLiquid}
    \frac{\mathrm{d}}{\mathrm{d}t}
    {\begin{bmatrix}
        \beta\\
        {\dot{\beta}}
    \end{bmatrix}}
    =
    \begin{bmatrix}
        \dot{\beta}\\
        f_{\beta}(\beta,\dot{\beta},\ddot{x}_c,\ddot{z}_c, \theta_c, \dot{\theta}_c, \ddot{\theta}_c)
    \end{bmatrix}.
\end{equation}
The parameters $d$ and $m$ of the model depend on the properties of the liquid and the container and can be identified through experiments.

As long as the mass does not experience lateral acceleration in the local coordinate frame, the pendulum remains in the fixed point $\beta = 0$, see e.g. \cite{Muchacho.2022} for this assumption.

%% file: 04_Framework.tex
\section{Assistive Telemanipulation Framework based on MPC}
\label{sec:Framework}
In this section, our assistive telemanipulation framework is introduced. The goal of the human operator is to move the liquid container without spilling it. For our Assistive Telemanipulation Framework, we define the two objectives 
\begin{itemize}
    \item $\mathrm{O}_1$: tracking the given input and
    \item $\mathrm{O}_2$: stabilizing the liquid.
\end{itemize}

The proposed framework consists of three components: 1)~the two models from section \ref{section:models}, 2) the user interface with input mapping and human movement prediction, and 3) a MPC that combines the two aforementioned goals of the controller into a single objective function.


\subsection{User Interface: Input Mapping and Human Movement Prediction}

\begin{figure}
\begin{center}
\includegraphics{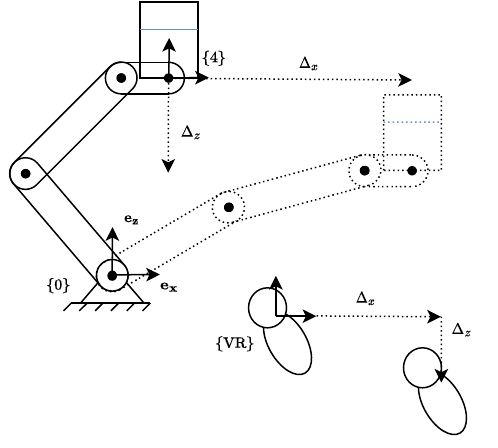}
\caption{Mapping of user input to desired end effector position. A movement to the right an downwards of the VR controller moves the desired position of the end effector in positive $\mathbf{e}_x$ and negative $\mathbf{e}_z$ direction.} 
\label{fig:UserInterface}
\end{center}
\end{figure}

The user interface for controlling the UR5e robot arm's end effector position is implemented using a \textit{Touch controller} for the \textit{Meta Quest 2} \cite{metaQuest2}. The operator can switch between an active and inactive mode using a button on the controller. In inactive mode, the desired position of the end effector is constantly set to the current position. When the mode switches to active, the current position of the end effector and the controller are saved as reference points. Relative movements of the controller are used as relative displacements of the desired end effector position, as shown in Fig. \ref{fig:UserInterface}. Controller movements to the right and left are mapped to the $\mathbf{e}_x$ axis, and movements up and down are mapped to the $\mathbf{e}_z$ axis, which is called position-position control \cite{Siciliano.2016}. 

To implement a MPC, predicting the human input over the prediction horizon $h_p$ is necessary. As suggested in \cite{Chipalkatty.2013}, the challenges of predicting human input due to the complexity and uncertainty of human behavior, can be overcame by the use of a simple prediction model. It has been shown that the proposed simple prediction model can reach sufficiently good results. In our MPC, the future reference positions are predicted assuming constant velocities.


\subsection{MPC for Assistive Telemanipulation}
The system states from the previously derived state space models (\ref{eq:stateSpaceModelRobot}) and        (\ref{eq:stateSpaceModelLiquid}) can be combined to
        \begin{align}
        \label{eq:combinedSystem}
            \dot{\mathbf{\xi}} = \frac{\mathrm{d}}{\mathrm{d}t} \begin{bmatrix}
                \mathbf{q}\\
                \mathbf{\dot{q}}\\
                \beta\\
                \dot{\beta}
            \end{bmatrix} = \mathbf{f_{\xi}}(\mathbf{\xi},\mathbf{u})
        \end{align}
         with dimension $n_{\xi} = 8$. Note, that the first 6 states are given in the joint space and the dynamics of the slosh angle $\beta$ are stated in the task space, thus the forward kinematics (\ref{eq:ForwardKinematics}) and its derivatives with respect to time are required.

System (\ref{eq:combinedSystem}) is discretized in time with the sampling period $\Delta t$ using Euler Forward, which is sufficient for this use case if the sampling period is sufficiently small. Assuming zero-order hold on the input $\mathbf{u}$, the discretized system results in  
\begin{equation}
    \mathbf{\xi}_{k+1} = \mathbf{\xi}_k + \mathbf{f}_{\mathbf{\xi},k} \cdot \Delta t.
\end{equation}

For the control task $\mathrm{O}_1$, we define the following tracking error function
\begin{align}
    \mathbf{e} = \begin{bmatrix}
        x_c\\
        z_c\\
        \theta_c
    \end{bmatrix} - 
    \begin{bmatrix}
        x_c\\
        z_c\\
        \theta_c    
    \end{bmatrix}_{ref}.
\end{align}
Since the system dynamics of the robot (\ref{eq:RoboterDynamik}) only contain the joint angles and the tracking error is described in the task space, the forward kinematics (\ref{eq:ForwardKinematics}) are implicitly used. 

The two control objectives $\mathrm{O}_1$ and $\mathrm{O}_2$ can now be combined into the discrete finite horizon objective function
\begin{align} \label{eq:objectiveFunction}
    J = \sum_{k=0}^{N-1} ||\mathbf{e}_{k+1}||^{2}_{\mathbf{Q_1}} +  ||\beta_{k+1}||^{2}_{Q_2}  +  ||\mathbf{u}_k||^{2}_{\mathbf{R}} 
\end{align}
with no stage cost of $e$ and $\beta$ at $k=0$ and $u_k$ at $k=N$. $\mathbf{Q_1} = \mathrm{diag}(Q_{1,1}, Q_{1,2}, Q_{1,3})$ is a diagonal matrix containing the weights for the tracking error in $x_c, z_c$ and $\theta_c$. The sloshing angle $\beta$ is weighted with $Q_2$ and stabilizing weights $\mathbf{R} = \mathrm{diag}(R_1,R_2,R_3)$ are added to prevent high accelerations on the joints, especially if the robot configuration is close to singularities.

Including system constraints and input constraints as well as starting states $\xi_0$, we obtain the nonlinear optimization problem
\begin{align}
    &\underset{\mathbf{\xi}_{1 \rightarrow N},\mathbf{u}_{0 \rightarrow N-1}}{\min}  && J(\mathbf{\xi}_{1 \rightarrow N},\mathbf{u}_{0 \rightarrow N-1})  \nonumber\\
    &\text{s.t.} \nonumber \\
    &&& \mathbf{\xi}_{k+1} = \mathbf{\xi}_k +  \mathbf{f}_{\xi,k}  \cdot \Delta t  \nonumber\\
    &&& \mathbf{q}_{\min} \leq \mathbf{q_k} \leq \mathbf{q}_{\max} \quad \forall k \in  [1, N] \nonumber\\
    &&& \mathbf{\dot{q}}_{\min} \leq \mathbf{\dot{q}_k} \leq \mathbf{\dot{q}}_{\max} \quad \forall k \in  [1, N] \nonumber\\
    &&& \mathbf{u}_{\min} \leq \mathbf{u_k} \leq \mathbf{u}_{\max} \quad \forall k \in  [0, N-1] \nonumber\\
    &&& \xi_{k=0} = \xi_0 \nonumber.
\end{align}
The inequality constraints on $\mathbf{q_k}$ for the robot joints are set according to the robot manual. Constraints on the joint velocities $\mathbf{\dot{q}_k}$ and acceleration $\mathbf{u_k}$ can be utilized to ensure slower robot movements for safety reasons. Note that through the usage of the forward kinematics, no inverse kinematics need to be solved after planning trajectories in the task space (\!\!\cite{Fau:17}, \cite{Rubagotti.2019}).
The optimization is solved with multiple shooting, thus the optimization variables include the input $\mathbf{u}$ as well as the system states $\mathbf{\xi}$. 

\begin{figure*}
\begin{center}
\includegraphics{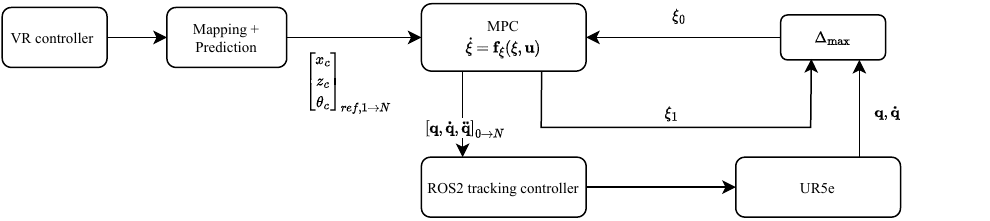}   
\caption{Overview of the control architecture.} 
\label{fig:Regelkreis}
\end{center}
\end{figure*}

In Fig. \ref{fig:Regelkreis} the overall control architecture is depicted. The states $\beta$ and $\dot{\beta}$ are not observable in the current setup and thus controlled in open loop.

With the assumption, that the underlying tracking controller runs at a high frequency with good tracking performance and assumption, that the solution trajectories coming from the optimization satisfy the system dynamics, also the measurable states $\mathbf{q}$ and $\mathbf{\dot{q}}$ are only fed back if the difference between predicted and measured states exceeds the threshold $\Delta_{max}$ to increase stability \cite[chap. 3]{Werling.2017}.


\subsection{Setup of the Assistive Telemanipulation Framework}
The two control objectives $\mathrm{O}_1$ and $\mathrm{O}_2$ are inherently contradictory, and therefore, the weights of the objective function need to be chosen such that a suitable compromise is achieved. The selection of the objective function weights is crucial and has a significant impact on the overall performance. 

With our Assistive Telemanipulation Framework, the operator can either use the remote robot with pure tracking behavior by prioritizing $\mathrm{O}_1$, or receive high support with stabilizing a liquid through a high prioritization of $\mathrm{O}_2$.

In this regard, multiobjective optimization can be a useful tool for identifying optimal weights. However, incorporating human factors in the optimization process can be challenging, and the trade-off between performance and user satisfaction must be carefully considered.

%% file: 05_Experiments.tex
\section{Realization and Evaluation of Assistive Telemanipulation Framework}
\label{sec:Realization}
This section presents the real-world realization of the framework and the illustrative evaluation with a human operator.

\subsection{Technical Setup of the System}
The implementation of the proposed method was carried out using ROS2 with version Foxy \cite{Macenski.2022}. For solving the optimization problem, the framework \textit{CasADi} \cite{Andersson.2019} with the solver \textit{ipopt} was used and ran with an update rate of $30 \mathrm{Hz}$. The framework was executed on a host PC with an Intel i7 8700 processor. The ROS2 control trajectory tracking controller was used to control the motion of the UE5e robot. The lengths $L_1 = 0.425\, \mathrm{m}$, $L_2 = 0.3922\, \mathrm{m}$ and $L_3 = 0.1\, \mathrm{m}$ were taken from the UR5e manual. The liquid model parameters, including the pendulum lengths $l=0.02$, mass $m=1$, damping coefficient $d=0.005$ and pendulum height $h=0.08$, were roughly identified through video analysis of the glass with water (also compare \cite{Biagiotti.2018} and \cite{Mayer.2012}).


\subsection{Experiment and Results}

We conducted an experiment with two parametrizations $\mathrm{P}_1$ and $\mathrm{P}_2$ (Tab. \ref{tab:objectiveWeights}) of the objective function (\ref{eq:objectiveFunction}) to validate our Assistive Telemanipulation Framework and investigate the influence of the contrary control objectives $\mathrm{O}_1$ and $\mathrm{O}_2$. To ensure comparability, we used the same starting position and recorded one human input with a length of 10 seconds, that was used with both parameter sets.

\begin{table}[htbp]
    \centering
    \caption{Weights of the objective function for testing the two different control objectives.}
    \label{tab:objectiveWeights}
    \begin{tabular}{|c|c|c|}
        \hline
        & $\mathrm{P}_1$: high tracking & $\mathrm{P}_2$: anti slosh \\
        \hline
        $Q_{1,1}$ & 500 & 100 \\
        $Q_{1,2}$ & 500 & 100 \\
        $Q_{1,3}$ & 100 & 1 \\
        $Q_2$ & 0.1 & 1000 \\
        $R_1$ & 0.01 & 0.01 \\
        $R_2$ & 0.01 & 0.01 \\
        $R_3$ & 0.01 & 0.01 \\
        \hline
    \end{tabular}
\end{table}

\begin{figure}[h]
\begin{center}
\input{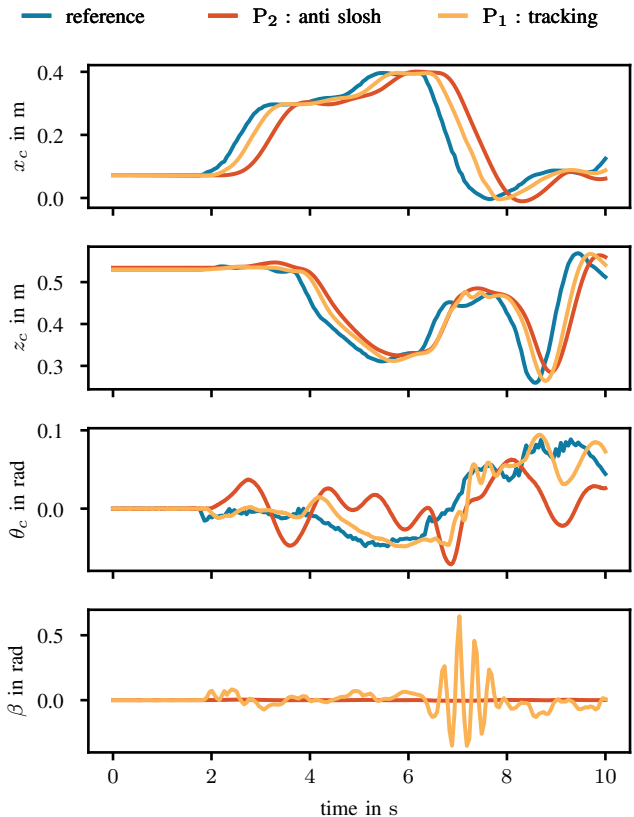}
\caption{Plots of the two parametrizations $\mathrm{P}_1$ and $\mathrm{P}_2$. Depicted are the reference and the actual trajectories in the task space as well as the calculated slosh angle $\beta$.}
\label{fig:PlotResults}
\end{center}
\end{figure}

Our qualitative results showed that the Assistive Telemanipulation Framework achieved a high tracking rate with sloshing motion when using $\mathrm{P}_1$, while $\mathrm{P}_2$ resulted in no sloshing but a higher delay. The quantitative results confirm this observation, as the high tracking rate almost perfectly follows the given input for $\mathrm{P}_1$, although some deviation is induced through the velocity constraints in the MPC as can be seen in Fig. \ref{fig:PlotResults}. The data of $\mathrm{P}_2$ shows a higher delay and a deviation in $\theta$ can be observed as the controller stabilizes the liquid. 

We also plotted the calculated sloshing angle $\beta$ in the same figure \ref{fig:PlotResults} to visualize the sloshing motion and the effectiveness of the Assistive Telemanipulation Framework to support the operator with stabilizing the liquid. With $\mathrm{P}_2$, basically no amplitude of the sloshing motion is achieved, whereas $\mathrm{P}_1$ induces high sloshing dynamics. 

\begin{figure}[h]
\begin{center}
\input{Execution_Time.pgf}
\caption{Execution times of the two parametrizations $\mathrm{P}_1$ and $\mathrm{P}_2$.}
\label{fig:MPCTimes}
\end{center}
\end{figure}
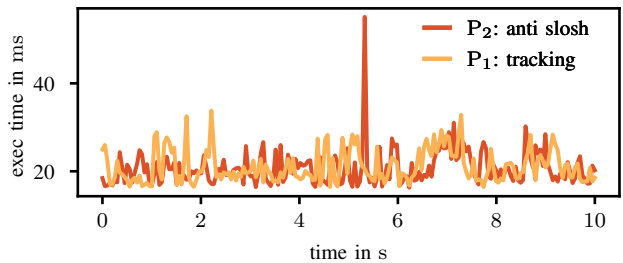

In addition to the results, we also measured the calculation time of the MPC algorithm over time which can be seen in Fig. \ref{fig:MPCTimes}. The sample rate is sufficient with a mean around $t_{\mathrm{mean}}= 21 \,\mathrm{ms}$ for real-time application of the Assistive Telemanipulation Framework. Regarding \cite{Andersson.2019}, the calculation time can be further reduced by a factor of 10.

Overall, our experiment demonstrates the trade-offs between the contrary control objectives $\mathrm{O}_1$ and $\mathrm{O}_2$ and the effective support of the human operator while stabilizing the liquid through our Assistive Telemanipulation Framework.

%% file: Execution_Time.pgf
\begingroup%
\makeatletter%
\begin{pgfpicture}%
\pgfpathrectangle{\pgfpointorigin}{\pgfqpoint{3.429748in}{1.544894in}}%
\pgfusepath{use as bounding box, clip}%
\begin{pgfscope}%
\pgfsetbuttcap%
\pgfsetmiterjoin%
\definecolor{currentfill}{rgb}{1.000000,1.000000,1.000000}%
\pgfsetfillcolor{currentfill}%
\pgfsetlinewidth{0.000000pt}%
\definecolor{currentstroke}{rgb}{1.000000,1.000000,1.000000}%
\pgfsetstrokecolor{currentstroke}%
\pgfsetdash{}{0pt}%
\pgfpathmoveto{\pgfqpoint{0.000000in}{0.000000in}}%
\pgfpathlineto{\pgfqpoint{3.429748in}{0.000000in}}%
\pgfpathlineto{\pgfqpoint{3.429748in}{1.544894in}}%
\pgfpathlineto{\pgfqpoint{0.000000in}{1.544894in}}%
\pgfpathlineto{\pgfqpoint{0.000000in}{0.000000in}}%
\pgfpathclose%
\pgfusepath{fill}%
\end{pgfscope}%
\begin{pgfscope}%
\pgfsetbuttcap%
\pgfsetmiterjoin%
\definecolor{currentfill}{rgb}{1.000000,1.000000,1.000000}%
\pgfsetfillcolor{currentfill}%
\pgfsetlinewidth{0.000000pt}%
\definecolor{currentstroke}{rgb}{0.000000,0.000000,0.000000}%
\pgfsetstrokecolor{currentstroke}%
\pgfsetstrokeopacity{0.000000}%
\pgfsetdash{}{0pt}%
\pgfpathmoveto{\pgfqpoint{0.494748in}{0.460894in}}%
\pgfpathlineto{\pgfqpoint{3.329748in}{0.460894in}}%
\pgfpathlineto{\pgfqpoint{3.329748in}{1.444894in}}%
\pgfpathlineto{\pgfqpoint{0.494748in}{1.444894in}}%
\pgfpathlineto{\pgfqpoint{0.494748in}{0.460894in}}%
\pgfpathclose%
\pgfusepath{fill}%
\end{pgfscope}%
\begin{pgfscope}%
\pgfsetbuttcap%
\pgfsetroundjoin%
\definecolor{currentfill}{rgb}{0.000000,0.000000,0.000000}%
\pgfsetfillcolor{currentfill}%
\pgfsetlinewidth{1.003750pt}%
\definecolor{currentstroke}{rgb}{0.000000,0.000000,0.000000}%
\pgfsetstrokecolor{currentstroke}%
\pgfsetdash{}{0pt}%
\pgfsys@defobject{currentmarker}{\pgfqpoint{0.000000in}{-0.041667in}}{\pgfqpoint{0.000000in}{0.000000in}}{%
\pgfpathmoveto{\pgfqpoint{0.000000in}{0.000000in}}%
\pgfpathlineto{\pgfqpoint{0.000000in}{-0.041667in}}%
\pgfusepath{stroke,fill}%
}%
\begin{pgfscope}%
\pgfsys@transformshift{0.623612in}{0.460894in}%
\pgfsys@useobject{currentmarker}{}%
\end{pgfscope}%
\end{pgfscope}%
\begin{pgfscope}%
\definecolor{textcolor}{rgb}{0.000000,0.000000,0.000000}%
\pgfsetstrokecolor{textcolor}%
\pgfsetfillcolor{textcolor}%
\pgftext[x=0.623612in,y=0.370616in,,top]{\color{textcolor}\rmfamily\fontsize{8.000000}{9.600000}\selectfont \(\displaystyle {0}\)}%
\end{pgfscope}%
\begin{pgfscope}%
\pgfsetbuttcap%
\pgfsetroundjoin%
\definecolor{currentfill}{rgb}{0.000000,0.000000,0.000000}%
\pgfsetfillcolor{currentfill}%
\pgfsetlinewidth{1.003750pt}%
\definecolor{currentstroke}{rgb}{0.000000,0.000000,0.000000}%
\pgfsetstrokecolor{currentstroke}%
\pgfsetdash{}{0pt}%
\pgfsys@defobject{currentmarker}{\pgfqpoint{0.000000in}{-0.041667in}}{\pgfqpoint{0.000000in}{0.000000in}}{%
\pgfpathmoveto{\pgfqpoint{0.000000in}{0.000000in}}%
\pgfpathlineto{\pgfqpoint{0.000000in}{-0.041667in}}%
\pgfusepath{stroke,fill}%
}%
\begin{pgfscope}%
\pgfsys@transformshift{1.139066in}{0.460894in}%
\pgfsys@useobject{currentmarker}{}%
\end{pgfscope}%
\end{pgfscope}%
\begin{pgfscope}%
\definecolor{textcolor}{rgb}{0.000000,0.000000,0.000000}%
\pgfsetstrokecolor{textcolor}%
\pgfsetfillcolor{textcolor}%
\pgftext[x=1.139066in,y=0.370616in,,top]{\color{textcolor}\rmfamily\fontsize{8.000000}{9.600000}\selectfont \(\displaystyle {2}\)}%
\end{pgfscope}%
\begin{pgfscope}%
\pgfsetbuttcap%
\pgfsetroundjoin%
\definecolor{currentfill}{rgb}{0.000000,0.000000,0.000000}%
\pgfsetfillcolor{currentfill}%
\pgfsetlinewidth{1.003750pt}%
\definecolor{currentstroke}{rgb}{0.000000,0.000000,0.000000}%
\pgfsetstrokecolor{currentstroke}%
\pgfsetdash{}{0pt}%
\pgfsys@defobject{currentmarker}{\pgfqpoint{0.000000in}{-0.041667in}}{\pgfqpoint{0.000000in}{0.000000in}}{%
\pgfpathmoveto{\pgfqpoint{0.000000in}{0.000000in}}%
\pgfpathlineto{\pgfqpoint{0.000000in}{-0.041667in}}%
\pgfusepath{stroke,fill}%
}%
\begin{pgfscope}%
\pgfsys@transformshift{1.654521in}{0.460894in}%
\pgfsys@useobject{currentmarker}{}%
\end{pgfscope}%
\end{pgfscope}%
\begin{pgfscope}%
\definecolor{textcolor}{rgb}{0.000000,0.000000,0.000000}%
\pgfsetstrokecolor{textcolor}%
\pgfsetfillcolor{textcolor}%
\pgftext[x=1.654521in,y=0.370616in,,top]{\color{textcolor}\rmfamily\fontsize{8.000000}{9.600000}\selectfont \(\displaystyle {4}\)}%
\end{pgfscope}%
\begin{pgfscope}%
\pgfsetbuttcap%
\pgfsetroundjoin%
\definecolor{currentfill}{rgb}{0.000000,0.000000,0.000000}%
\pgfsetfillcolor{currentfill}%
\pgfsetlinewidth{1.003750pt}%
\definecolor{currentstroke}{rgb}{0.000000,0.000000,0.000000}%
\pgfsetstrokecolor{currentstroke}%
\pgfsetdash{}{0pt}%
\pgfsys@defobject{currentmarker}{\pgfqpoint{0.000000in}{-0.041667in}}{\pgfqpoint{0.000000in}{0.000000in}}{%
\pgfpathmoveto{\pgfqpoint{0.000000in}{0.000000in}}%
\pgfpathlineto{\pgfqpoint{0.000000in}{-0.041667in}}%
\pgfusepath{stroke,fill}%
}%
\begin{pgfscope}%
\pgfsys@transformshift{2.169975in}{0.460894in}%
\pgfsys@useobject{currentmarker}{}%
\end{pgfscope}%
\end{pgfscope}%
\begin{pgfscope}%
\definecolor{textcolor}{rgb}{0.000000,0.000000,0.000000}%
\pgfsetstrokecolor{textcolor}%
\pgfsetfillcolor{textcolor}%
\pgftext[x=2.169975in,y=0.370616in,,top]{\color{textcolor}\rmfamily\fontsize{8.000000}{9.600000}\selectfont \(\displaystyle {6}\)}%
\end{pgfscope}%
\begin{pgfscope}%
\pgfsetbuttcap%
\pgfsetroundjoin%
\definecolor{currentfill}{rgb}{0.000000,0.000000,0.000000}%
\pgfsetfillcolor{currentfill}%
\pgfsetlinewidth{1.003750pt}%
\definecolor{currentstroke}{rgb}{0.000000,0.000000,0.000000}%
\pgfsetstrokecolor{currentstroke}%
\pgfsetdash{}{0pt}%
\pgfsys@defobject{currentmarker}{\pgfqpoint{0.000000in}{-0.041667in}}{\pgfqpoint{0.000000in}{0.000000in}}{%
\pgfpathmoveto{\pgfqpoint{0.000000in}{0.000000in}}%
\pgfpathlineto{\pgfqpoint{0.000000in}{-0.041667in}}%
\pgfusepath{stroke,fill}%
}%
\begin{pgfscope}%
\pgfsys@transformshift{2.685430in}{0.460894in}%
\pgfsys@useobject{currentmarker}{}%
\end{pgfscope}%
\end{pgfscope}%
\begin{pgfscope}%
\definecolor{textcolor}{rgb}{0.000000,0.000000,0.000000}%
\pgfsetstrokecolor{textcolor}%
\pgfsetfillcolor{textcolor}%
\pgftext[x=2.685430in,y=0.370616in,,top]{\color{textcolor}\rmfamily\fontsize{8.000000}{9.600000}\selectfont \(\displaystyle {8}\)}%
\end{pgfscope}%
\begin{pgfscope}%
\pgfsetbuttcap%
\pgfsetroundjoin%
\definecolor{currentfill}{rgb}{0.000000,0.000000,0.000000}%
\pgfsetfillcolor{currentfill}%
\pgfsetlinewidth{1.003750pt}%
\definecolor{currentstroke}{rgb}{0.000000,0.000000,0.000000}%
\pgfsetstrokecolor{currentstroke}%
\pgfsetdash{}{0pt}%
\pgfsys@defobject{currentmarker}{\pgfqpoint{0.000000in}{-0.041667in}}{\pgfqpoint{0.000000in}{0.000000in}}{%
\pgfpathmoveto{\pgfqpoint{0.000000in}{0.000000in}}%
\pgfpathlineto{\pgfqpoint{0.000000in}{-0.041667in}}%
\pgfusepath{stroke,fill}%
}%
\begin{pgfscope}%
\pgfsys@transformshift{3.200885in}{0.460894in}%
\pgfsys@useobject{currentmarker}{}%
\end{pgfscope}%
\end{pgfscope}%
\begin{pgfscope}%
\definecolor{textcolor}{rgb}{0.000000,0.000000,0.000000}%
\pgfsetstrokecolor{textcolor}%
\pgfsetfillcolor{textcolor}%
\pgftext[x=3.200885in,y=0.370616in,,top]{\color{textcolor}\rmfamily\fontsize{8.000000}{9.600000}\selectfont \(\displaystyle {10}\)}%
\end{pgfscope}%
\begin{pgfscope}%
\definecolor{textcolor}{rgb}{0.000000,0.000000,0.000000}%
\pgfsetstrokecolor{textcolor}%
\pgfsetfillcolor{textcolor}%
\pgftext[x=1.912248in,y=0.207530in,,top]{\color{textcolor}\rmfamily\fontsize{8.000000}{9.600000}\selectfont time in \(\displaystyle \mathrm{s}\)}%
\end{pgfscope}%
\begin{pgfscope}%
\pgfsetbuttcap%
\pgfsetroundjoin%
\definecolor{currentfill}{rgb}{0.000000,0.000000,0.000000}%
\pgfsetfillcolor{currentfill}%
\pgfsetlinewidth{1.003750pt}%
\definecolor{currentstroke}{rgb}{0.000000,0.000000,0.000000}%
\pgfsetstrokecolor{currentstroke}%
\pgfsetdash{}{0pt}%
\pgfsys@defobject{currentmarker}{\pgfqpoint{-0.041667in}{0.000000in}}{\pgfqpoint{-0.000000in}{0.000000in}}{%
\pgfpathmoveto{\pgfqpoint{-0.000000in}{0.000000in}}%
\pgfpathlineto{\pgfqpoint{-0.041667in}{0.000000in}}%
\pgfusepath{stroke,fill}%
}%
\begin{pgfscope}%
\pgfsys@transformshift{0.494748in}{0.591490in}%
\pgfsys@useobject{currentmarker}{}%
\end{pgfscope}%
\end{pgfscope}%
\begin{pgfscope}%
\definecolor{textcolor}{rgb}{0.000000,0.000000,0.000000}%
\pgfsetstrokecolor{textcolor}%
\pgfsetfillcolor{textcolor}%
\pgftext[x=0.263086in, y=0.549281in, left, base]{\color{textcolor}\rmfamily\fontsize{8.000000}{9.600000}\selectfont 20}%
\end{pgfscope}%
\begin{pgfscope}%
\pgfsetbuttcap%
\pgfsetroundjoin%
\definecolor{currentfill}{rgb}{0.000000,0.000000,0.000000}%
\pgfsetfillcolor{currentfill}%
\pgfsetlinewidth{1.003750pt}%
\definecolor{currentstroke}{rgb}{0.000000,0.000000,0.000000}%
\pgfsetstrokecolor{currentstroke}%
\pgfsetdash{}{0pt}%
\pgfsys@defobject{currentmarker}{\pgfqpoint{-0.041667in}{0.000000in}}{\pgfqpoint{-0.000000in}{0.000000in}}{%
\pgfpathmoveto{\pgfqpoint{-0.000000in}{0.000000in}}%
\pgfpathlineto{\pgfqpoint{-0.041667in}{0.000000in}}%
\pgfusepath{stroke,fill}%
}%
\begin{pgfscope}%
\pgfsys@transformshift{0.494748in}{1.052007in}%
\pgfsys@useobject{currentmarker}{}%
\end{pgfscope}%
\end{pgfscope}%
\begin{pgfscope}%
\definecolor{textcolor}{rgb}{0.000000,0.000000,0.000000}%
\pgfsetstrokecolor{textcolor}%
\pgfsetfillcolor{textcolor}%
\pgftext[x=0.263086in, y=1.009798in, left, base]{\color{textcolor}\rmfamily\fontsize{8.000000}{9.600000}\selectfont 40}%
\end{pgfscope}%
\begin{pgfscope}%
\definecolor{textcolor}{rgb}{0.000000,0.000000,0.000000}%
\pgfsetstrokecolor{textcolor}%
\pgfsetfillcolor{textcolor}%
\pgftext[x=0.207530in,y=0.952894in,,bottom,rotate=90.000000]{\color{textcolor}\rmfamily\fontsize{8.000000}{9.600000}\selectfont exec time in  \(\displaystyle \mathrm{ms}\) }%
\end{pgfscope}%
\begin{pgfscope}%
\pgfpathrectangle{\pgfqpoint{0.494748in}{0.460894in}}{\pgfqpoint{2.835000in}{0.984000in}}%
\pgfusepath{clip}%
\pgfsetrectcap%
\pgfsetroundjoin%
\pgfsetlinewidth{1.505625pt}%
\definecolor{currentstroke}{rgb}{0.866667,0.317647,0.160784}%
\pgfsetstrokecolor{currentstroke}%
\pgfsetdash{}{0pt}%
\pgfpathmoveto{\pgfqpoint{0.623612in}{0.555864in}}%
\pgfpathlineto{\pgfqpoint{0.636563in}{0.513177in}}%
\pgfpathlineto{\pgfqpoint{0.649514in}{0.516789in}}%
\pgfpathlineto{\pgfqpoint{0.662465in}{0.528909in}}%
\pgfpathlineto{\pgfqpoint{0.675416in}{0.517072in}}%
\pgfpathlineto{\pgfqpoint{0.688367in}{0.539171in}}%
\pgfpathlineto{\pgfqpoint{0.701319in}{0.533850in}}%
\pgfpathlineto{\pgfqpoint{0.714270in}{0.691544in}}%
\pgfpathlineto{\pgfqpoint{0.727221in}{0.612066in}}%
\pgfpathlineto{\pgfqpoint{0.740172in}{0.529155in}}%
\pgfpathlineto{\pgfqpoint{0.753123in}{0.626155in}}%
\pgfpathlineto{\pgfqpoint{0.766074in}{0.603496in}}%
\pgfpathlineto{\pgfqpoint{0.779025in}{0.531396in}}%
\pgfpathlineto{\pgfqpoint{0.791976in}{0.581664in}}%
\pgfpathlineto{\pgfqpoint{0.804928in}{0.629748in}}%
\pgfpathlineto{\pgfqpoint{0.817879in}{0.701970in}}%
\pgfpathlineto{\pgfqpoint{0.830830in}{0.681763in}}%
\pgfpathlineto{\pgfqpoint{0.843781in}{0.585303in}}%
\pgfpathlineto{\pgfqpoint{0.856732in}{0.628198in}}%
\pgfpathlineto{\pgfqpoint{0.869683in}{0.527636in}}%
\pgfpathlineto{\pgfqpoint{0.882634in}{0.518956in}}%
\pgfpathlineto{\pgfqpoint{0.895585in}{0.578421in}}%
\pgfpathlineto{\pgfqpoint{0.908536in}{0.640927in}}%
\pgfpathlineto{\pgfqpoint{0.921488in}{0.519896in}}%
\pgfpathlineto{\pgfqpoint{0.934439in}{0.533052in}}%
\pgfpathlineto{\pgfqpoint{0.947390in}{0.632512in}}%
\pgfpathlineto{\pgfqpoint{0.960341in}{0.668049in}}%
\pgfpathlineto{\pgfqpoint{0.973292in}{0.602441in}}%
\pgfpathlineto{\pgfqpoint{0.986243in}{0.602263in}}%
\pgfpathlineto{\pgfqpoint{0.999194in}{0.637762in}}%
\pgfpathlineto{\pgfqpoint{1.012145in}{0.520678in}}%
\pgfpathlineto{\pgfqpoint{1.025097in}{0.635328in}}%
\pgfpathlineto{\pgfqpoint{1.038048in}{0.535631in}}%
\pgfpathlineto{\pgfqpoint{1.050999in}{0.570145in}}%
\pgfpathlineto{\pgfqpoint{1.063950in}{0.546923in}}%
\pgfpathlineto{\pgfqpoint{1.076901in}{0.553073in}}%
\pgfpathlineto{\pgfqpoint{1.089852in}{0.591391in}}%
\pgfpathlineto{\pgfqpoint{1.102803in}{0.580921in}}%
\pgfpathlineto{\pgfqpoint{1.115754in}{0.596119in}}%
\pgfpathlineto{\pgfqpoint{1.128705in}{0.577678in}}%
\pgfpathlineto{\pgfqpoint{1.141657in}{0.581994in}}%
\pgfpathlineto{\pgfqpoint{1.154608in}{0.675679in}}%
\pgfpathlineto{\pgfqpoint{1.167559in}{0.615825in}}%
\pgfpathlineto{\pgfqpoint{1.180510in}{0.542865in}}%
\pgfpathlineto{\pgfqpoint{1.193461in}{0.525682in}}%
\pgfpathlineto{\pgfqpoint{1.206412in}{0.523949in}}%
\pgfpathlineto{\pgfqpoint{1.219363in}{0.571959in}}%
\pgfpathlineto{\pgfqpoint{1.232314in}{0.604433in}}%
\pgfpathlineto{\pgfqpoint{1.245266in}{0.590744in}}%
\pgfpathlineto{\pgfqpoint{1.258217in}{0.652174in}}%
\pgfpathlineto{\pgfqpoint{1.271168in}{0.570994in}}%
\pgfpathlineto{\pgfqpoint{1.284119in}{0.577578in}}%
\pgfpathlineto{\pgfqpoint{1.297070in}{0.571413in}}%
\pgfpathlineto{\pgfqpoint{1.310021in}{0.598875in}}%
\pgfpathlineto{\pgfqpoint{1.322972in}{0.606632in}}%
\pgfpathlineto{\pgfqpoint{1.335923in}{0.610441in}}%
\pgfpathlineto{\pgfqpoint{1.348875in}{0.594176in}}%
\pgfpathlineto{\pgfqpoint{1.361826in}{0.521689in}}%
\pgfpathlineto{\pgfqpoint{1.374777in}{0.723609in}}%
\pgfpathlineto{\pgfqpoint{1.387728in}{0.580953in}}%
\pgfpathlineto{\pgfqpoint{1.400679in}{0.564158in}}%
\pgfpathlineto{\pgfqpoint{1.413630in}{0.568754in}}%
\pgfpathlineto{\pgfqpoint{1.426581in}{0.508230in}}%
\pgfpathlineto{\pgfqpoint{1.439532in}{0.579108in}}%
\pgfpathlineto{\pgfqpoint{1.452483in}{0.685393in}}%
\pgfpathlineto{\pgfqpoint{1.465435in}{0.744037in}}%
\pgfpathlineto{\pgfqpoint{1.478386in}{0.553177in}}%
\pgfpathlineto{\pgfqpoint{1.491337in}{0.532316in}}%
\pgfpathlineto{\pgfqpoint{1.504288in}{0.514062in}}%
\pgfpathlineto{\pgfqpoint{1.517239in}{0.552702in}}%
\pgfpathlineto{\pgfqpoint{1.530190in}{0.705206in}}%
\pgfpathlineto{\pgfqpoint{1.543141in}{0.520399in}}%
\pgfpathlineto{\pgfqpoint{1.556092in}{0.728933in}}%
\pgfpathlineto{\pgfqpoint{1.569044in}{0.534987in}}%
\pgfpathlineto{\pgfqpoint{1.581995in}{0.558030in}}%
\pgfpathlineto{\pgfqpoint{1.594946in}{0.678881in}}%
\pgfpathlineto{\pgfqpoint{1.607897in}{0.630032in}}%
\pgfpathlineto{\pgfqpoint{1.620848in}{0.578609in}}%
\pgfpathlineto{\pgfqpoint{1.633799in}{0.656176in}}%
\pgfpathlineto{\pgfqpoint{1.646750in}{0.607665in}}%
\pgfpathlineto{\pgfqpoint{1.659701in}{0.635117in}}%
\pgfpathlineto{\pgfqpoint{1.672652in}{0.620592in}}%
\pgfpathlineto{\pgfqpoint{1.685604in}{0.624558in}}%
\pgfpathlineto{\pgfqpoint{1.698555in}{0.644571in}}%
\pgfpathlineto{\pgfqpoint{1.711506in}{0.649314in}}%
\pgfpathlineto{\pgfqpoint{1.724457in}{0.528146in}}%
\pgfpathlineto{\pgfqpoint{1.737408in}{0.509544in}}%
\pgfpathlineto{\pgfqpoint{1.750359in}{0.590011in}}%
\pgfpathlineto{\pgfqpoint{1.763310in}{0.665683in}}%
\pgfpathlineto{\pgfqpoint{1.776261in}{0.541430in}}%
\pgfpathlineto{\pgfqpoint{1.789213in}{0.536100in}}%
\pgfpathlineto{\pgfqpoint{1.802164in}{0.520670in}}%
\pgfpathlineto{\pgfqpoint{1.815115in}{0.612458in}}%
\pgfpathlineto{\pgfqpoint{1.828066in}{0.640656in}}%
\pgfpathlineto{\pgfqpoint{1.841017in}{0.674235in}}%
\pgfpathlineto{\pgfqpoint{1.853968in}{0.637703in}}%
\pgfpathlineto{\pgfqpoint{1.866919in}{0.521911in}}%
\pgfpathlineto{\pgfqpoint{1.879870in}{0.529230in}}%
\pgfpathlineto{\pgfqpoint{1.892822in}{0.609226in}}%
\pgfpathlineto{\pgfqpoint{1.905773in}{0.711779in}}%
\pgfpathlineto{\pgfqpoint{1.918724in}{0.544925in}}%
\pgfpathlineto{\pgfqpoint{1.931675in}{0.509919in}}%
\pgfpathlineto{\pgfqpoint{1.944626in}{0.637921in}}%
\pgfpathlineto{\pgfqpoint{1.957577in}{0.511237in}}%
\pgfpathlineto{\pgfqpoint{1.970528in}{0.535454in}}%
\pgfpathlineto{\pgfqpoint{1.983479in}{0.632428in}}%
\pgfpathlineto{\pgfqpoint{1.996430in}{1.400167in}}%
\pgfpathlineto{\pgfqpoint{2.009382in}{0.603227in}}%
\pgfpathlineto{\pgfqpoint{2.022333in}{0.532885in}}%
\pgfpathlineto{\pgfqpoint{2.035284in}{0.557813in}}%
\pgfpathlineto{\pgfqpoint{2.048235in}{0.511579in}}%
\pgfpathlineto{\pgfqpoint{2.061186in}{0.555979in}}%
\pgfpathlineto{\pgfqpoint{2.074137in}{0.719493in}}%
\pgfpathlineto{\pgfqpoint{2.087088in}{0.511083in}}%
\pgfpathlineto{\pgfqpoint{2.100039in}{0.550566in}}%
\pgfpathlineto{\pgfqpoint{2.112991in}{0.623875in}}%
\pgfpathlineto{\pgfqpoint{2.125942in}{0.549352in}}%
\pgfpathlineto{\pgfqpoint{2.138893in}{0.763528in}}%
\pgfpathlineto{\pgfqpoint{2.151844in}{0.721584in}}%
\pgfpathlineto{\pgfqpoint{2.164795in}{0.732042in}}%
\pgfpathlineto{\pgfqpoint{2.177746in}{0.515676in}}%
\pgfpathlineto{\pgfqpoint{2.190697in}{0.530144in}}%
\pgfpathlineto{\pgfqpoint{2.203648in}{0.556652in}}%
\pgfpathlineto{\pgfqpoint{2.216599in}{0.565448in}}%
\pgfpathlineto{\pgfqpoint{2.229551in}{0.505621in}}%
\pgfpathlineto{\pgfqpoint{2.242502in}{0.635022in}}%
\pgfpathlineto{\pgfqpoint{2.255453in}{0.542196in}}%
\pgfpathlineto{\pgfqpoint{2.268404in}{0.547550in}}%
\pgfpathlineto{\pgfqpoint{2.281355in}{0.588725in}}%
\pgfpathlineto{\pgfqpoint{2.294306in}{0.543577in}}%
\pgfpathlineto{\pgfqpoint{2.307257in}{0.573029in}}%
\pgfpathlineto{\pgfqpoint{2.320208in}{0.561392in}}%
\pgfpathlineto{\pgfqpoint{2.333160in}{0.621347in}}%
\pgfpathlineto{\pgfqpoint{2.346111in}{0.563151in}}%
\pgfpathlineto{\pgfqpoint{2.359062in}{0.648321in}}%
\pgfpathlineto{\pgfqpoint{2.372013in}{0.694514in}}%
\pgfpathlineto{\pgfqpoint{2.384964in}{0.699251in}}%
\pgfpathlineto{\pgfqpoint{2.397915in}{0.740698in}}%
\pgfpathlineto{\pgfqpoint{2.410866in}{0.711043in}}%
\pgfpathlineto{\pgfqpoint{2.423817in}{0.714078in}}%
\pgfpathlineto{\pgfqpoint{2.436769in}{0.796311in}}%
\pgfpathlineto{\pgfqpoint{2.449720in}{0.655893in}}%
\pgfpathlineto{\pgfqpoint{2.462671in}{0.845768in}}%
\pgfpathlineto{\pgfqpoint{2.475622in}{0.649793in}}%
\pgfpathlineto{\pgfqpoint{2.488573in}{0.665080in}}%
\pgfpathlineto{\pgfqpoint{2.501524in}{0.757026in}}%
\pgfpathlineto{\pgfqpoint{2.514475in}{0.738200in}}%
\pgfpathlineto{\pgfqpoint{2.527426in}{0.696214in}}%
\pgfpathlineto{\pgfqpoint{2.540377in}{0.720791in}}%
\pgfpathlineto{\pgfqpoint{2.553329in}{0.640856in}}%
\pgfpathlineto{\pgfqpoint{2.566280in}{0.586585in}}%
\pgfpathlineto{\pgfqpoint{2.579231in}{0.548030in}}%
\pgfpathlineto{\pgfqpoint{2.592182in}{0.522022in}}%
\pgfpathlineto{\pgfqpoint{2.605133in}{0.694520in}}%
\pgfpathlineto{\pgfqpoint{2.618084in}{0.744818in}}%
\pgfpathlineto{\pgfqpoint{2.631035in}{0.712539in}}%
\pgfpathlineto{\pgfqpoint{2.643986in}{0.565068in}}%
\pgfpathlineto{\pgfqpoint{2.656938in}{0.552124in}}%
\pgfpathlineto{\pgfqpoint{2.669889in}{0.627095in}}%
\pgfpathlineto{\pgfqpoint{2.682840in}{0.598739in}}%
\pgfpathlineto{\pgfqpoint{2.695791in}{0.561622in}}%
\pgfpathlineto{\pgfqpoint{2.708742in}{0.601901in}}%
\pgfpathlineto{\pgfqpoint{2.721693in}{0.558399in}}%
\pgfpathlineto{\pgfqpoint{2.734644in}{0.603568in}}%
\pgfpathlineto{\pgfqpoint{2.747595in}{0.611141in}}%
\pgfpathlineto{\pgfqpoint{2.760547in}{0.567931in}}%
\pgfpathlineto{\pgfqpoint{2.773498in}{0.540513in}}%
\pgfpathlineto{\pgfqpoint{2.786449in}{0.615267in}}%
\pgfpathlineto{\pgfqpoint{2.799400in}{0.544107in}}%
\pgfpathlineto{\pgfqpoint{2.812351in}{0.531283in}}%
\pgfpathlineto{\pgfqpoint{2.825302in}{0.545333in}}%
\pgfpathlineto{\pgfqpoint{2.838253in}{0.826559in}}%
\pgfpathlineto{\pgfqpoint{2.851204in}{0.737192in}}%
\pgfpathlineto{\pgfqpoint{2.864155in}{0.658840in}}%
\pgfpathlineto{\pgfqpoint{2.877107in}{0.610868in}}%
\pgfpathlineto{\pgfqpoint{2.890058in}{0.727594in}}%
\pgfpathlineto{\pgfqpoint{2.903009in}{0.682826in}}%
\pgfpathlineto{\pgfqpoint{2.915960in}{0.705635in}}%
\pgfpathlineto{\pgfqpoint{2.928911in}{0.619008in}}%
\pgfpathlineto{\pgfqpoint{2.941862in}{0.785824in}}%
\pgfpathlineto{\pgfqpoint{2.954813in}{0.629506in}}%
\pgfpathlineto{\pgfqpoint{2.967764in}{0.589575in}}%
\pgfpathlineto{\pgfqpoint{2.980716in}{0.544282in}}%
\pgfpathlineto{\pgfqpoint{2.993667in}{0.564763in}}%
\pgfpathlineto{\pgfqpoint{3.006618in}{0.537293in}}%
\pgfpathlineto{\pgfqpoint{3.019569in}{0.580826in}}%
\pgfpathlineto{\pgfqpoint{3.032520in}{0.562444in}}%
\pgfpathlineto{\pgfqpoint{3.045471in}{0.549238in}}%
\pgfpathlineto{\pgfqpoint{3.058422in}{0.639282in}}%
\pgfpathlineto{\pgfqpoint{3.071373in}{0.591182in}}%
\pgfpathlineto{\pgfqpoint{3.084324in}{0.529576in}}%
\pgfpathlineto{\pgfqpoint{3.097276in}{0.553903in}}%
\pgfpathlineto{\pgfqpoint{3.110227in}{0.634415in}}%
\pgfpathlineto{\pgfqpoint{3.123178in}{0.670088in}}%
\pgfpathlineto{\pgfqpoint{3.136129in}{0.569249in}}%
\pgfpathlineto{\pgfqpoint{3.149080in}{0.551491in}}%
\pgfpathlineto{\pgfqpoint{3.162031in}{0.526750in}}%
\pgfpathlineto{\pgfqpoint{3.174982in}{0.527733in}}%
\pgfpathlineto{\pgfqpoint{3.187933in}{0.621332in}}%
\pgfpathlineto{\pgfqpoint{3.200885in}{0.596627in}}%
\pgfpathlineto{\pgfqpoint{3.200885in}{0.596627in}}%
\pgfusepath{stroke}%
\end{pgfscope}%
\begin{pgfscope}%
\pgfpathrectangle{\pgfqpoint{0.494748in}{0.460894in}}{\pgfqpoint{2.835000in}{0.984000in}}%
\pgfusepath{clip}%
\pgfsetrectcap%
\pgfsetroundjoin%
\pgfsetlinewidth{1.505625pt}%
\definecolor{currentstroke}{rgb}{0.980392,0.698039,0.333333}%
\pgfsetstrokecolor{currentstroke}%
\pgfsetdash{}{0pt}%
\pgfpathmoveto{\pgfqpoint{0.623612in}{0.706354in}}%
\pgfpathlineto{\pgfqpoint{0.636563in}{0.729149in}}%
\pgfpathlineto{\pgfqpoint{0.649514in}{0.646496in}}%
\pgfpathlineto{\pgfqpoint{0.662465in}{0.524733in}}%
\pgfpathlineto{\pgfqpoint{0.675416in}{0.521128in}}%
\pgfpathlineto{\pgfqpoint{0.688367in}{0.560647in}}%
\pgfpathlineto{\pgfqpoint{0.701319in}{0.646703in}}%
\pgfpathlineto{\pgfqpoint{0.714270in}{0.583263in}}%
\pgfpathlineto{\pgfqpoint{0.727221in}{0.576090in}}%
\pgfpathlineto{\pgfqpoint{0.740172in}{0.560671in}}%
\pgfpathlineto{\pgfqpoint{0.753123in}{0.544134in}}%
\pgfpathlineto{\pgfqpoint{0.766074in}{0.510417in}}%
\pgfpathlineto{\pgfqpoint{0.779025in}{0.546231in}}%
\pgfpathlineto{\pgfqpoint{0.791976in}{0.557178in}}%
\pgfpathlineto{\pgfqpoint{0.804928in}{0.531963in}}%
\pgfpathlineto{\pgfqpoint{0.817879in}{0.569274in}}%
\pgfpathlineto{\pgfqpoint{0.830830in}{0.513652in}}%
\pgfpathlineto{\pgfqpoint{0.843781in}{0.520966in}}%
\pgfpathlineto{\pgfqpoint{0.856732in}{0.530736in}}%
\pgfpathlineto{\pgfqpoint{0.869683in}{0.513438in}}%
\pgfpathlineto{\pgfqpoint{0.882634in}{0.512484in}}%
\pgfpathlineto{\pgfqpoint{0.895585in}{0.776285in}}%
\pgfpathlineto{\pgfqpoint{0.908536in}{0.796066in}}%
\pgfpathlineto{\pgfqpoint{0.921488in}{0.595956in}}%
\pgfpathlineto{\pgfqpoint{0.934439in}{0.548235in}}%
\pgfpathlineto{\pgfqpoint{0.947390in}{0.738192in}}%
\pgfpathlineto{\pgfqpoint{0.960341in}{0.768994in}}%
\pgfpathlineto{\pgfqpoint{0.973292in}{0.744349in}}%
\pgfpathlineto{\pgfqpoint{0.986243in}{0.688764in}}%
\pgfpathlineto{\pgfqpoint{0.999194in}{0.716216in}}%
\pgfpathlineto{\pgfqpoint{1.012145in}{0.521572in}}%
\pgfpathlineto{\pgfqpoint{1.025097in}{0.567025in}}%
\pgfpathlineto{\pgfqpoint{1.038048in}{0.521903in}}%
\pgfpathlineto{\pgfqpoint{1.050999in}{0.634738in}}%
\pgfpathlineto{\pgfqpoint{1.063950in}{0.879953in}}%
\pgfpathlineto{\pgfqpoint{1.076901in}{0.555286in}}%
\pgfpathlineto{\pgfqpoint{1.089852in}{0.523835in}}%
\pgfpathlineto{\pgfqpoint{1.102803in}{0.509838in}}%
\pgfpathlineto{\pgfqpoint{1.115754in}{0.546267in}}%
\pgfpathlineto{\pgfqpoint{1.128705in}{0.589164in}}%
\pgfpathlineto{\pgfqpoint{1.141657in}{0.588974in}}%
\pgfpathlineto{\pgfqpoint{1.154608in}{0.618050in}}%
\pgfpathlineto{\pgfqpoint{1.167559in}{0.702489in}}%
\pgfpathlineto{\pgfqpoint{1.180510in}{0.665959in}}%
\pgfpathlineto{\pgfqpoint{1.193461in}{0.908503in}}%
\pgfpathlineto{\pgfqpoint{1.206412in}{0.717837in}}%
\pgfpathlineto{\pgfqpoint{1.219363in}{0.561798in}}%
\pgfpathlineto{\pgfqpoint{1.232314in}{0.575960in}}%
\pgfpathlineto{\pgfqpoint{1.245266in}{0.548797in}}%
\pgfpathlineto{\pgfqpoint{1.258217in}{0.585777in}}%
\pgfpathlineto{\pgfqpoint{1.271168in}{0.574718in}}%
\pgfpathlineto{\pgfqpoint{1.284119in}{0.585675in}}%
\pgfpathlineto{\pgfqpoint{1.297070in}{0.548713in}}%
\pgfpathlineto{\pgfqpoint{1.310021in}{0.589270in}}%
\pgfpathlineto{\pgfqpoint{1.322972in}{0.542259in}}%
\pgfpathlineto{\pgfqpoint{1.335923in}{0.580680in}}%
\pgfpathlineto{\pgfqpoint{1.348875in}{0.562534in}}%
\pgfpathlineto{\pgfqpoint{1.361826in}{0.614588in}}%
\pgfpathlineto{\pgfqpoint{1.374777in}{0.579761in}}%
\pgfpathlineto{\pgfqpoint{1.387728in}{0.586279in}}%
\pgfpathlineto{\pgfqpoint{1.400679in}{0.594899in}}%
\pgfpathlineto{\pgfqpoint{1.413630in}{0.647460in}}%
\pgfpathlineto{\pgfqpoint{1.426581in}{0.560060in}}%
\pgfpathlineto{\pgfqpoint{1.439532in}{0.590967in}}%
\pgfpathlineto{\pgfqpoint{1.452483in}{0.656278in}}%
\pgfpathlineto{\pgfqpoint{1.465435in}{0.533074in}}%
\pgfpathlineto{\pgfqpoint{1.478386in}{0.626280in}}%
\pgfpathlineto{\pgfqpoint{1.491337in}{0.593041in}}%
\pgfpathlineto{\pgfqpoint{1.504288in}{0.579664in}}%
\pgfpathlineto{\pgfqpoint{1.517239in}{0.537079in}}%
\pgfpathlineto{\pgfqpoint{1.530190in}{0.542408in}}%
\pgfpathlineto{\pgfqpoint{1.543141in}{0.534213in}}%
\pgfpathlineto{\pgfqpoint{1.556092in}{0.585363in}}%
\pgfpathlineto{\pgfqpoint{1.569044in}{0.558023in}}%
\pgfpathlineto{\pgfqpoint{1.594946in}{0.656073in}}%
\pgfpathlineto{\pgfqpoint{1.607897in}{0.582242in}}%
\pgfpathlineto{\pgfqpoint{1.620848in}{0.659550in}}%
\pgfpathlineto{\pgfqpoint{1.633799in}{0.565661in}}%
\pgfpathlineto{\pgfqpoint{1.646750in}{0.561636in}}%
\pgfpathlineto{\pgfqpoint{1.659701in}{0.548886in}}%
\pgfpathlineto{\pgfqpoint{1.672652in}{0.556963in}}%
\pgfpathlineto{\pgfqpoint{1.685604in}{0.592896in}}%
\pgfpathlineto{\pgfqpoint{1.698555in}{0.575933in}}%
\pgfpathlineto{\pgfqpoint{1.711506in}{0.572637in}}%
\pgfpathlineto{\pgfqpoint{1.724457in}{0.544780in}}%
\pgfpathlineto{\pgfqpoint{1.737408in}{0.547642in}}%
\pgfpathlineto{\pgfqpoint{1.750359in}{0.762065in}}%
\pgfpathlineto{\pgfqpoint{1.763310in}{0.516523in}}%
\pgfpathlineto{\pgfqpoint{1.776261in}{0.571579in}}%
\pgfpathlineto{\pgfqpoint{1.789213in}{0.757863in}}%
\pgfpathlineto{\pgfqpoint{1.802164in}{0.782075in}}%
\pgfpathlineto{\pgfqpoint{1.815115in}{0.508870in}}%
\pgfpathlineto{\pgfqpoint{1.828066in}{0.660599in}}%
\pgfpathlineto{\pgfqpoint{1.841017in}{0.531821in}}%
\pgfpathlineto{\pgfqpoint{1.853968in}{0.511376in}}%
\pgfpathlineto{\pgfqpoint{1.866919in}{0.619656in}}%
\pgfpathlineto{\pgfqpoint{1.879870in}{0.766957in}}%
\pgfpathlineto{\pgfqpoint{1.905773in}{0.579259in}}%
\pgfpathlineto{\pgfqpoint{1.918724in}{0.710426in}}%
\pgfpathlineto{\pgfqpoint{1.931675in}{0.784287in}}%
\pgfpathlineto{\pgfqpoint{1.944626in}{0.731990in}}%
\pgfpathlineto{\pgfqpoint{1.957577in}{0.775782in}}%
\pgfpathlineto{\pgfqpoint{1.970528in}{0.656943in}}%
\pgfpathlineto{\pgfqpoint{1.983479in}{0.615863in}}%
\pgfpathlineto{\pgfqpoint{1.996430in}{0.605043in}}%
\pgfpathlineto{\pgfqpoint{2.009382in}{0.573167in}}%
\pgfpathlineto{\pgfqpoint{2.022333in}{0.568560in}}%
\pgfpathlineto{\pgfqpoint{2.035284in}{0.543765in}}%
\pgfpathlineto{\pgfqpoint{2.048235in}{0.539240in}}%
\pgfpathlineto{\pgfqpoint{2.061186in}{0.720581in}}%
\pgfpathlineto{\pgfqpoint{2.074137in}{0.553241in}}%
\pgfpathlineto{\pgfqpoint{2.087088in}{0.577935in}}%
\pgfpathlineto{\pgfqpoint{2.100039in}{0.528745in}}%
\pgfpathlineto{\pgfqpoint{2.112991in}{0.555952in}}%
\pgfpathlineto{\pgfqpoint{2.125942in}{0.613956in}}%
\pgfpathlineto{\pgfqpoint{2.138893in}{0.664251in}}%
\pgfpathlineto{\pgfqpoint{2.151844in}{0.650465in}}%
\pgfpathlineto{\pgfqpoint{2.164795in}{0.638252in}}%
\pgfpathlineto{\pgfqpoint{2.177746in}{0.514684in}}%
\pgfpathlineto{\pgfqpoint{2.190697in}{0.581553in}}%
\pgfpathlineto{\pgfqpoint{2.203648in}{0.574343in}}%
\pgfpathlineto{\pgfqpoint{2.216599in}{0.557163in}}%
\pgfpathlineto{\pgfqpoint{2.229551in}{0.516235in}}%
\pgfpathlineto{\pgfqpoint{2.242502in}{0.559209in}}%
\pgfpathlineto{\pgfqpoint{2.255453in}{0.572862in}}%
\pgfpathlineto{\pgfqpoint{2.268404in}{0.620263in}}%
\pgfpathlineto{\pgfqpoint{2.281355in}{0.711602in}}%
\pgfpathlineto{\pgfqpoint{2.294306in}{0.598075in}}%
\pgfpathlineto{\pgfqpoint{2.307257in}{0.722259in}}%
\pgfpathlineto{\pgfqpoint{2.320208in}{0.760950in}}%
\pgfpathlineto{\pgfqpoint{2.333160in}{0.701415in}}%
\pgfpathlineto{\pgfqpoint{2.346111in}{0.671336in}}%
\pgfpathlineto{\pgfqpoint{2.359062in}{0.785295in}}%
\pgfpathlineto{\pgfqpoint{2.372013in}{0.690820in}}%
\pgfpathlineto{\pgfqpoint{2.384964in}{0.733715in}}%
\pgfpathlineto{\pgfqpoint{2.397915in}{0.807386in}}%
\pgfpathlineto{\pgfqpoint{2.410866in}{0.721551in}}%
\pgfpathlineto{\pgfqpoint{2.423817in}{0.779541in}}%
\pgfpathlineto{\pgfqpoint{2.436769in}{0.770821in}}%
\pgfpathlineto{\pgfqpoint{2.449720in}{0.672848in}}%
\pgfpathlineto{\pgfqpoint{2.462671in}{0.671559in}}%
\pgfpathlineto{\pgfqpoint{2.475622in}{0.667080in}}%
\pgfpathlineto{\pgfqpoint{2.488573in}{0.791670in}}%
\pgfpathlineto{\pgfqpoint{2.501524in}{0.887267in}}%
\pgfpathlineto{\pgfqpoint{2.514475in}{0.634946in}}%
\pgfpathlineto{\pgfqpoint{2.527426in}{0.589765in}}%
\pgfpathlineto{\pgfqpoint{2.540377in}{0.554463in}}%
\pgfpathlineto{\pgfqpoint{2.553329in}{0.571982in}}%
\pgfpathlineto{\pgfqpoint{2.566280in}{0.658327in}}%
\pgfpathlineto{\pgfqpoint{2.579231in}{0.540219in}}%
\pgfpathlineto{\pgfqpoint{2.592182in}{0.549084in}}%
\pgfpathlineto{\pgfqpoint{2.605133in}{0.545646in}}%
\pgfpathlineto{\pgfqpoint{2.618084in}{0.508997in}}%
\pgfpathlineto{\pgfqpoint{2.631035in}{0.556545in}}%
\pgfpathlineto{\pgfqpoint{2.643986in}{0.597263in}}%
\pgfpathlineto{\pgfqpoint{2.656938in}{0.709051in}}%
\pgfpathlineto{\pgfqpoint{2.669889in}{0.612404in}}%
\pgfpathlineto{\pgfqpoint{2.682840in}{0.541217in}}%
\pgfpathlineto{\pgfqpoint{2.695791in}{0.531732in}}%
\pgfpathlineto{\pgfqpoint{2.708742in}{0.552922in}}%
\pgfpathlineto{\pgfqpoint{2.721693in}{0.543508in}}%
\pgfpathlineto{\pgfqpoint{2.734644in}{0.627998in}}%
\pgfpathlineto{\pgfqpoint{2.747595in}{0.631451in}}%
\pgfpathlineto{\pgfqpoint{2.760547in}{0.636384in}}%
\pgfpathlineto{\pgfqpoint{2.773498in}{0.618670in}}%
\pgfpathlineto{\pgfqpoint{2.786449in}{0.557721in}}%
\pgfpathlineto{\pgfqpoint{2.799400in}{0.552638in}}%
\pgfpathlineto{\pgfqpoint{2.812351in}{0.582427in}}%
\pgfpathlineto{\pgfqpoint{2.825302in}{0.586588in}}%
\pgfpathlineto{\pgfqpoint{2.838253in}{0.649326in}}%
\pgfpathlineto{\pgfqpoint{2.851204in}{0.690675in}}%
\pgfpathlineto{\pgfqpoint{2.864155in}{0.782398in}}%
\pgfpathlineto{\pgfqpoint{2.877107in}{0.622850in}}%
\pgfpathlineto{\pgfqpoint{2.890058in}{0.543889in}}%
\pgfpathlineto{\pgfqpoint{2.903009in}{0.654761in}}%
\pgfpathlineto{\pgfqpoint{2.915960in}{0.676725in}}%
\pgfpathlineto{\pgfqpoint{2.928911in}{0.646679in}}%
\pgfpathlineto{\pgfqpoint{2.941862in}{0.612077in}}%
\pgfpathlineto{\pgfqpoint{2.954813in}{0.625365in}}%
\pgfpathlineto{\pgfqpoint{2.967764in}{0.707918in}}%
\pgfpathlineto{\pgfqpoint{2.980716in}{0.783854in}}%
\pgfpathlineto{\pgfqpoint{2.993667in}{0.734038in}}%
\pgfpathlineto{\pgfqpoint{3.006618in}{0.701389in}}%
\pgfpathlineto{\pgfqpoint{3.019569in}{0.607084in}}%
\pgfpathlineto{\pgfqpoint{3.032520in}{0.556613in}}%
\pgfpathlineto{\pgfqpoint{3.045471in}{0.623204in}}%
\pgfpathlineto{\pgfqpoint{3.058422in}{0.632066in}}%
\pgfpathlineto{\pgfqpoint{3.071373in}{0.558257in}}%
\pgfpathlineto{\pgfqpoint{3.084324in}{0.559301in}}%
\pgfpathlineto{\pgfqpoint{3.097276in}{0.611761in}}%
\pgfpathlineto{\pgfqpoint{3.110227in}{0.632630in}}%
\pgfpathlineto{\pgfqpoint{3.123178in}{0.552254in}}%
\pgfpathlineto{\pgfqpoint{3.136129in}{0.534068in}}%
\pgfpathlineto{\pgfqpoint{3.149080in}{0.538244in}}%
\pgfpathlineto{\pgfqpoint{3.162031in}{0.538462in}}%
\pgfpathlineto{\pgfqpoint{3.174982in}{0.609702in}}%
\pgfpathlineto{\pgfqpoint{3.187933in}{0.537899in}}%
\pgfpathlineto{\pgfqpoint{3.200885in}{0.558069in}}%
\pgfpathlineto{\pgfqpoint{3.200885in}{0.558069in}}%
\pgfusepath{stroke}%
\end{pgfscope}%
\begin{pgfscope}%
\pgfsetrectcap%
\pgfsetmiterjoin%
\pgfsetlinewidth{1.003750pt}%
\definecolor{currentstroke}{rgb}{0.000000,0.000000,0.000000}%
\pgfsetstrokecolor{currentstroke}%
\pgfsetdash{}{0pt}%
\pgfpathmoveto{\pgfqpoint{0.494748in}{0.460894in}}%
\pgfpathlineto{\pgfqpoint{0.494748in}{1.444894in}}%
\pgfusepath{stroke}%
\end{pgfscope}%
\begin{pgfscope}%
\pgfsetrectcap%
\pgfsetmiterjoin%
\pgfsetlinewidth{1.003750pt}%
\definecolor{currentstroke}{rgb}{0.000000,0.000000,0.000000}%
\pgfsetstrokecolor{currentstroke}%
\pgfsetdash{}{0pt}%
\pgfpathmoveto{\pgfqpoint{3.329748in}{0.460894in}}%
\pgfpathlineto{\pgfqpoint{3.329748in}{1.444894in}}%
\pgfusepath{stroke}%
\end{pgfscope}%
\begin{pgfscope}%
\pgfsetrectcap%
\pgfsetmiterjoin%
\pgfsetlinewidth{1.003750pt}%
\definecolor{currentstroke}{rgb}{0.000000,0.000000,0.000000}%
\pgfsetstrokecolor{currentstroke}%
\pgfsetdash{}{0pt}%
\pgfpathmoveto{\pgfqpoint{0.494748in}{0.460894in}}%
\pgfpathlineto{\pgfqpoint{3.329748in}{0.460894in}}%
\pgfusepath{stroke}%
\end{pgfscope}%
\begin{pgfscope}%
\pgfsetrectcap%
\pgfsetmiterjoin%
\pgfsetlinewidth{1.003750pt}%
\definecolor{currentstroke}{rgb}{0.000000,0.000000,0.000000}%
\pgfsetstrokecolor{currentstroke}%
\pgfsetdash{}{0pt}%
\pgfpathmoveto{\pgfqpoint{0.494748in}{1.444894in}}%
\pgfpathlineto{\pgfqpoint{3.329748in}{1.444894in}}%
\pgfusepath{stroke}%
\end{pgfscope}%
\begin{pgfscope}%
\pgfsetrectcap%
\pgfsetroundjoin%
\pgfsetlinewidth{1.505625pt}%
\definecolor{currentstroke}{rgb}{0.866667,0.317647,0.160784}%
\pgfsetstrokecolor{currentstroke}%
\pgfsetdash{}{0pt}%
\pgfpathmoveto{\pgfqpoint{2.309259in}{1.343809in}}%
\pgfpathlineto{\pgfqpoint{2.375926in}{1.343809in}}%
\pgfpathlineto{\pgfqpoint{2.442593in}{1.343809in}}%
\pgfusepath{stroke}%
\end{pgfscope}%
\begin{pgfscope}%
\definecolor{textcolor}{rgb}{0.000000,0.000000,0.000000}%
\pgfsetstrokecolor{textcolor}%
\pgfsetfillcolor{textcolor}%
\pgftext[x=2.531482in,y=1.304920in,left,base]{\color{textcolor}\rmfamily\fontsize{8.000000}{9.600000}\selectfont \(\displaystyle \mathrm{P}_2\): anti slosh}%
\end{pgfscope}%
\begin{pgfscope}%
\pgfsetrectcap%
\pgfsetroundjoin%
\pgfsetlinewidth{1.505625pt}%
\definecolor{currentstroke}{rgb}{0.980392,0.698039,0.333333}%
\pgfsetstrokecolor{currentstroke}%
\pgfsetdash{}{0pt}%
\pgfpathmoveto{\pgfqpoint{2.309259in}{1.180723in}}%
\pgfpathlineto{\pgfqpoint{2.375926in}{1.180723in}}%
\pgfpathlineto{\pgfqpoint{2.442593in}{1.180723in}}%
\pgfusepath{stroke}%
\end{pgfscope}%
\begin{pgfscope}%
\definecolor{textcolor}{rgb}{0.000000,0.000000,0.000000}%
\pgfsetstrokecolor{textcolor}%
\pgfsetfillcolor{textcolor}%
\pgftext[x=2.531482in,y=1.141834in,left,base]{\color{textcolor}\rmfamily\fontsize{8.000000}{9.600000}\selectfont \(\displaystyle \mathrm{P}_1\): tracking}%
\end{pgfscope}%
\begin{pgfscope}%
\pgfsetrectcap%
\pgfsetroundjoin%
\pgfsetlinewidth{1.505625pt}%
\definecolor{currentstroke}{rgb}{0.866667,0.317647,0.160784}%
\pgfsetstrokecolor{currentstroke}%
\pgfsetdash{}{0pt}%
\pgfpathmoveto{\pgfqpoint{2.309259in}{1.343809in}}%
\pgfpathlineto{\pgfqpoint{2.375926in}{1.343809in}}%
\pgfpathlineto{\pgfqpoint{2.442593in}{1.343809in}}%
\pgfusepath{stroke}%
\end{pgfscope}%
\begin{pgfscope}%
\definecolor{textcolor}{rgb}{0.000000,0.000000,0.000000}%
\pgfsetstrokecolor{textcolor}%
\pgfsetfillcolor{textcolor}%
\pgftext[x=2.531482in,y=1.304920in,left,base]{\color{textcolor}\rmfamily\fontsize{8.000000}{9.600000}\selectfont \(\displaystyle \mathrm{P}_2\): anti slosh}%
\end{pgfscope}%
\begin{pgfscope}%
\pgfsetrectcap%
\pgfsetroundjoin%
\pgfsetlinewidth{1.505625pt}%
\definecolor{currentstroke}{rgb}{0.980392,0.698039,0.333333}%
\pgfsetstrokecolor{currentstroke}%
\pgfsetdash{}{0pt}%
\pgfpathmoveto{\pgfqpoint{2.309259in}{1.180723in}}%
\pgfpathlineto{\pgfqpoint{2.375926in}{1.180723in}}%
\pgfpathlineto{\pgfqpoint{2.442593in}{1.180723in}}%
\pgfusepath{stroke}%
\end{pgfscope}%
\begin{pgfscope}%
\definecolor{textcolor}{rgb}{0.000000,0.000000,0.000000}%
\pgfsetstrokecolor{textcolor}%
\pgfsetfillcolor{textcolor}%
\pgftext[x=2.531482in,y=1.141834in,left,base]{\color{textcolor}\rmfamily\fontsize{8.000000}{9.600000}\selectfont \(\displaystyle \mathrm{P}_1\): tracking}%
\end{pgfscope}%
\begin{pgfscope}%
\pgfsetrectcap%
\pgfsetroundjoin%
\pgfsetlinewidth{1.505625pt}%
\definecolor{currentstroke}{rgb}{0.866667,0.317647,0.160784}%
\pgfsetstrokecolor{currentstroke}%
\pgfsetdash{}{0pt}%
\pgfpathmoveto{\pgfqpoint{2.309259in}{1.343809in}}%
\pgfpathlineto{\pgfqpoint{2.375926in}{1.343809in}}%
\pgfpathlineto{\pgfqpoint{2.442593in}{1.343809in}}%
\pgfusepath{stroke}%
\end{pgfscope}%
\begin{pgfscope}%
\definecolor{textcolor}{rgb}{0.000000,0.000000,0.000000}%
\pgfsetstrokecolor{textcolor}%
\pgfsetfillcolor{textcolor}%
\pgftext[x=2.531482in,y=1.304920in,left,base]{\color{textcolor}\rmfamily\fontsize{8.000000}{9.600000}\selectfont \(\displaystyle \mathrm{P}_2\): anti slosh}%
\end{pgfscope}%
\begin{pgfscope}%
\pgfsetrectcap%
\pgfsetroundjoin%
\pgfsetlinewidth{1.505625pt}%
\definecolor{currentstroke}{rgb}{0.980392,0.698039,0.333333}%
\pgfsetstrokecolor{currentstroke}%
\pgfsetdash{}{0pt}%
\pgfpathmoveto{\pgfqpoint{2.309259in}{1.180723in}}%
\pgfpathlineto{\pgfqpoint{2.375926in}{1.180723in}}%
\pgfpathlineto{\pgfqpoint{2.442593in}{1.180723in}}%
\pgfusepath{stroke}%
\end{pgfscope}%
\begin{pgfscope}%
\definecolor{textcolor}{rgb}{0.000000,0.000000,0.000000}%
\pgfsetstrokecolor{textcolor}%
\pgfsetfillcolor{textcolor}%
\pgftext[x=2.531482in,y=1.141834in,left,base]{\color{textcolor}\rmfamily\fontsize{8.000000}{9.600000}\selectfont \(\displaystyle \mathrm{P}_1\): tracking}%
\end{pgfscope}%
\end{pgfpicture}%
\makeatother%
\endgroup%

%% file: 06_Conclusion.tex
\section{Conclusion}
\label{sec:Conclusion}

In this paper, a novel framework for telemanipulation is proposed using non-linear model predictive control and a virtual reality input device. 
Our framework addresses challenges such as insufficient transparency, low immersion, and limited feedback to the human operator. Furthermore, we provide a model for human input prediction and a model for slosh dynamics control with a liquid container, increasing the accuracy of remote tasks. The experiments are conducted on an UR5e robot arm with a glass of water connected to the end effector. 
The results indicated the effectiveness of the proposed framework in controlling the remote robot. The combination of non-linear model predictive control and a virtual reality input device provided a more intuitive and efficient interface for the operator to control the remote robot.

In future research, we plan to extend the framework to three-dimensional environments and investigate additional assistance features such as collision avoidance and pouring assistance.



%% file: SMC2023.bbl
\begin{thebibliography}{10}
\providecommand{\url}[1]{#1}
\csname url@samestyle\endcsname
\providecommand{\newblock}{\relax}
\providecommand{\bibinfo}[2]{#2}
\providecommand{\BIBentrySTDinterwordspacing}{\spaceskip=0pt\relax}
\providecommand{\BIBentryALTinterwordstretchfactor}{4}
\providecommand{\BIBentryALTinterwordspacing}{\spaceskip=\fontdimen2\font plus
\BIBentryALTinterwordstretchfactor\fontdimen3\font minus
  \fontdimen4\font\relax}
\providecommand{\BIBforeignlanguage}[2]{{%
\expandafter\ifx\csname l@#1\endcsname\relax
\typeout{** WARNING: IEEEtran.bst: No hyphenation pattern has been}%
\typeout{** loaded for the language `#1'. Using the pattern for}%
\typeout{** the default language instead.}%
\else
\language=\csname l@#1\endcsname
\fi
#2}}
\providecommand{\BIBdecl}{\relax}
\BIBdecl

\bibitem{Sheridan.2016}
T.~B. Sheridan, ``Human-robot interaction: Status and challenges,'' \emph{Human
  Factors}, vol.~58, no.~4, pp. 525--532, 2016.

\bibitem{Hokayem.2006}
P.~F. Hokayem and M.~W. Spong, ``Bilateral teleoperation: An historical
  survey,'' \emph{Automatica}, vol.~42, no.~12, pp. 2035--2057, 2006.

\bibitem{Hirche.2012}
S.~Hirche and M.~Buss, ``Human-oriented control for haptic teleoperation,''
  \emph{Proceedings of the IEEE}, vol. 100, no.~3, pp. 623--647, 2012.

\bibitem{Wildenbeest.2013}
J.~G.~W. Wildenbeest, D.~A. Abbink, C.~J.~M. Heemskerk, F.~C.~T. {van der
  Helm}, and H.~Boessenkool, ``The impact of haptic feedback quality on the
  performance of teleoperated assembly tasks,'' \emph{IEEE Transactions on
  Haptics}, vol.~6, no.~2, pp. 242--252, 2013.

\bibitem{Hertkorn.2016}
K.~Hertkorn, \emph{Shared Grasping: a Combination of Telepresence and Grasp
  Planning}.\hskip 1em plus 0.5em minus 0.4em\relax {KIT Scientific
  Publishing}, 2016.

\bibitem{Nostadt.2020}
N.~Nostadt, D.~A. Abbink, O.~Christ, and P.~Beckerle, ``Embodiment, presence,
  and their intersections,'' \emph{ACM Transactions on Human-Robot
  Interaction}, vol.~9, no.~4, pp. 1--19, 2020.

\bibitem{sigma7}
\BIBentryALTinterwordspacing
``Sigma.7.'' [Online]. Available:
  \url{https://www.forcedimension.com/products/sigma}
\BIBentrySTDinterwordspacing

\bibitem{Fau:17}
T.~Faulwasser, T.~Weber, P.~Zometa, and R.~Findeisen, ``Implementation of
  nonlinear model predictive path-following control for an industrial robot,''
  \emph{IEEE Transactions on Control Systems Technology}, vol.~25, no.~4, pp.
  1505--1511, 2017.

\bibitem{Hu.2021}
S.~Hu, E.~Babaians, M.~Karimi, and E.~Steinbach, ``Nmpc-mp: Real-time nonlinear
  model predictive control for safe motion planning in manipulator
  teleoperation,'' in \emph{2021 IEEE/RSJ International Conference on
  Intelligent Robots and Systems (IROS)}.\hskip 1em plus 0.5em minus
  0.4em\relax IEEE, 2021.

\bibitem{Rubagotti.2019}
M.~Rubagotti, T.~Taunyazov, B.~Omarali, and A.~Shintemirov, ``Semi-autonomous
  robot teleoperation with obstacle avoidance via model predictive control,''
  \emph{IEEE Robotics and Automation Letters}, vol.~4, no.~3, pp. 2746--2753,
  2019.

\bibitem{Nubert.2020}
J.~Nubert, J.~Kohler, V.~Berenz, F.~Allg{\"o}wer, and S.~Trimpe, ``Safe and
  fast tracking on a robot manipulator: Robust mpc and neural network
  control,'' \emph{IEEE Robotics and Automation Letters}, vol.~5, no.~2, pp.
  3050--3057, 2020.

\bibitem{Guagliumi.2021}
L.~Guagliumi, A.~Berti, E.~Monti, and M.~Carricato, ``A simple model-based
  method for sloshing estimation in liquid transfer in automatic machines,''
  \emph{IEEE Access}, vol.~9, pp. 129\,347--129\,357, 2021.

\bibitem{Feddema.1997}
J.~T. Feddema, C.~R. Dohrmann, G.~G. Parker, R.~D. Robinett, V.~J. Romero, and
  D.~J. Schmitt, ``Control for slosh-free motion of an open container,''
  \emph{IEEE Control Systems Magazine}, vol.~17, no.~1, pp. 29--36, 1997.

\bibitem{Biagiotti.2018}
L.~Biagiotti, D.~Chiaravalli, L.~Moriello, and C.~Melchiorri, ``A plug-in
  feed-forward control for sloshing suppression in robotic teleoperation
  tasks,'' in \emph{2018 IEEE/RSJ International Conference on Intelligent
  Robots and Systems (IROS)}, 2018, pp. 5855--5860.

\bibitem{ReinholdJan.2019}
{Reinhold Jan}, {Amersdorfer Manuel}, and {Meurer Thomas}, ``A dynamic
  optimization approach for sloshing free transport of liquid filled containers
  using an industrial robot,'' in \emph{2019 IEEE/RSJ International Conference
  on Intelligent Robots and Systems (IROS)}, 2019, pp. 2336--2341.

\bibitem{Maderna.2018}
R.~Maderna, A.~Casalino, A.~M. Zanchettin, and P.~Rocco, ``Robotic handling of
  liquids with spilling avoidance: a constraint-based control approach,'' in
  \emph{2018 IEEE International Conference on Robotics and Automation (ICRA)},
  2018, pp. 7414--7420.

\bibitem{Muchacho.2022}
R.~I.~C. Muchacho, R.~Laha, L.~F. Figueredo, and S.~Haddadin, ``A solution to
  slosh-free robot trajectory optimization,'' in \emph{2022 IEEE/RSJ
  International Conference on Intelligent Robots and Systems (IROS)}.\hskip 1em
  plus 0.5em minus 0.4em\relax IEEE, 2022.

\bibitem{Lynch.2017}
K.~M. Lynch and F.~C. Park, \emph{Modern robotics: Mechanics, planning, and
  control}.\hskip 1em plus 0.5em minus 0.4em\relax Cambridge and New York, NY
  and Port Melbourne: {Cambridge University Press}, 2017.

\bibitem{metaQuest2}
\BIBentryALTinterwordspacing
``Meta quest 2.'' [Online]. Available:
  \url{https://www.meta.com/quest/products/quest-2/}
\BIBentrySTDinterwordspacing

\bibitem{Siciliano.2016}
B.~Siciliano and O.~Khatib, \emph{Springer Handbook of Robotics}.\hskip 1em
  plus 0.5em minus 0.4em\relax Cham: {Springer International Publishing}, 2016.

\bibitem{Chipalkatty.2013}
R.~Chipalkatty, G.~Droge, and M.~B. Egerstedt, ``Less is more: Mixed-initiative
  model-predictive control with human inputs,'' \emph{IEEE Transactions on
  Robotics}, vol.~29, no.~3, pp. 695--703, 2013.

\bibitem{Werling.2017}
M.~Werling, \emph{Optimale aktive Fahreingriffe: F{\"u}r Sicherheits- und
  Komfortsysteme in Fahrzeugen}.\hskip 1em plus 0.5em minus 0.4em\relax Berlin:
  {De Gruyter}, 2017.

\bibitem{Macenski.2022}
\BIBentryALTinterwordspacing
S.~Macenski, T.~Foote, B.~Gerkey, C.~Lalancette, and W.~Woodall, ``Robot
  operating system 2: Design, architecture, and uses in the wild,''
  \emph{Science Robotics}, vol.~7, no.~66, p. eabm6074, 2022. [Online].
  Available: \url{https://www.science.org/doi/abs/10.1126/scirobotics.abm6074}
\BIBentrySTDinterwordspacing

\bibitem{Andersson.2019}
J.~A.~E. Andersson, J.~Gillis, G.~Horn, J.~B. Rawlings, and M.~Diehl,
  ``{CasADi} -- {A} software framework for nonlinear optimization and optimal
  control,'' \emph{Mathematical Programming Computation}, vol.~11, no.~1, pp.
  1--36, 2019.

\bibitem{Mayer.2012}
H.~C. Mayer and R.~Krechetnikov, ``Walking with coffee: Why does it spill?''
  \emph{Physical Review E}, vol.~85, no.~4, p. 046117, 2012.

\end{thebibliography}
